\documentclass[review]{elsarticle}

\usepackage{lineno,hyperref}
\usepackage[USenglish]{babel}
\usepackage{graphicx}
\usepackage{mdframed}
\usepackage{latexsym}
\usepackage{python}
\usepackage{wrapfig}
\usepackage{epstopdf}
\usepackage{array}
\usepackage{multirow}
\usepackage{caption}
\usepackage{wrapfig}
\usepackage{subfig}
\usepackage{amssymb}
\usepackage[cmex10]{amsmath}
\usepackage{pgfplots}
\usepackage{algorithm}
\usepackage[noend]{algpseudocode}
\usepackage{amsfonts}
\usepackage{footmisc}
\usepackage{rotating}
\usepackage{xparse}
\usepackage{todonotes}
\usepackage{comment}
\usepackage{placeins}
\usepackage{enumitem}
\usepackage{marginnote}
\usepackage{colortbl}
\usepackage{bigstrut}
\DeclareGraphicsExtensions{.eps}

\usepackage{amsthm}

\definecolor{amethyst}{rgb}{0.6, 0.4, 0.8}
\definecolor{americanrose}{rgb}{0.8, 0.0, 0.8}
\definecolor{green}{rgb}{0.31, 0.78, 0.47}
\definecolor{yellow}{rgb}{0.89, 0.82, 0.04}
\definecolor{amber}{rgb}{0.0, 0.5, 0.0}

\journal{Engineering Applications of Artificial Intelligence}

\begin{document}
\newtheorem{definition}{Definition}
\begin{frontmatter}

\title{Predictive Process Model Monitoring Using Recurrent Neural Networks}

\author[address1]{Johannes~De~Smedt\corref{mycorrespondingauthor}}
\cortext[mycorrespondingauthor]{Corresponding author}
\ead{johannes.desmedt@kuleuven.be}
\author[address1]{Jochen~De~Weerdt}
\ead{jochen.deweerdt@kuleuven.be}

\address[address1]{Research Centre for Information Systems Engineering (LIRIS)\\
KU Leuven, Belgium}

\begin{abstract}
The field of predictive process monitoring focuses on case-level models to predict a single specific outcome such as a particular objective, (remaining) time, or next activity/remaining sequence.
Recently, a longer-horizon, model-wide approach has been proposed in the form of process model forecasting, which predicts the future state of a whole process model through the forecasting of all activity-to-activity relations at once using time series forecasting.

This paper introduces the concept of \emph{predictive process model monitoring} which sits in the middle of both predictive process monitoring and process model forecasting.
Concretely, by modelling a process model as a set of constraints being present between activities over time, we can capture more detailed information between activities compared to process model forecasting, while being compatible with typical predictive process monitoring objectives which are often expressed in the same language as these constraints.
To achieve this, Processes-As-Movies (PAM) is introduced, i.e., a novel technique capable of jointly mining and predicting declarative process constraints between activities in various windows of a process' execution.
PAM predicts what declarative rules hold for a trace (objective-based), which also supports the prediction of all constraints together as a process model (model-based).
Various recurrent neural network topologies inspired by video analysis tailored to temporal high-dimensional input are used to model the process model evolution with windows as time steps, including encoder-decoder long short-term memory networks, and convolutional long short-term memory networks. Results obtained over real-life event logs show that these topologies are effective in terms of predictive accuracy and precision.
\end{abstract}


\end{frontmatter}

\section{Introduction}
\label{sec:intro}
Predictive Process Monitoring (PPM) has known a recent surge of interest, fuelled by the growth of new machine learning applications.
Most notably, the introduction of recurrent neural networks facilitated a leap in performance for remaining time and next-in-sequence prediction \cite{DBLP:journals/dss/EvermannRF17,DBLP:conf/caise/TaxVRD17}.
Others have focused efforts on predicting the outcome of particular objectives of a process execution \cite{DBLP:conf/caise/MaggiFDG14,DBLP:journals/corr/TeinemaaDRM17}, e.g., the acceptance of a loan application, or the execution of a credit check before starting the review of a loan application.
Both are based on using process execution prefixes, i.e., historic evidence of a process typically captured in event logs.
The former targets activity labels and execution time as dependent variables while the latter whether particular objectives formulated as rules, often linear temporal logic (LTL) rules between activities and/or data variables, hold as dependent variable(s).
On the other hand, a recent paradigm shift was proposed to move towards process system-wide predictions through Process Model Forecasting (PMF) \cite{smedt2021process}.
By predicting activity-to-activity directly-follows occurrences using univariate time series techniques, a long-term forecast is obtained of the process model supporting the system.
This allows to perform more strategic analyses compared to predictive process monitoring, which is a more operational and essentially case-based paradigm, but results in a more coarsely-distributed predictive accuracy.

In this paper, Predictive Process Model Monitoring (PPMM) is introduced which sits in the middle of both paradigms.
It converts a trace into a feature space by dividing it into windows and mining them for a set of declarative constraints between activities expressed in LTL Declare rules \cite{pesic2006declarative}.
This creates an evolving image of the process model underpinning its behavior over the subsequent windows, which allows obtaining a more detailed prediction between activities compared to process model forecasting (simple directly-follows vs. a full set of constraints), while offering trace-based predictions in LTL which are similar to many of the objectives used in PPM.
Trace-based predictions are able to generate information such as which activities will be executed next (by their presence in particular constraints such as \texttt{existence} constraints), and what particular (control flow-based) LTL objectives have been met (such as \texttt{(chain) precedence} between different activities).
The predictions on a trace level can be aggregated to form a full process model as they are expressed in the same set of relations.
In the extreme case, with windows coinciding with single activities, PPMM can achieve similar results to regular prefix-based predictions.


Drawing inspiration from research on the analysis of moving images over time, where convolutional neural networks take care of the high dimensionality of the input stages (images or frames in a video) and long short-term memory networks (LSTMs) deal with the time evolution, Process-As-Movies (PAM) converts processes into a similar structure that treats windows within traces as high-dimensional inputs.
More specifically, we obtain a high-dimensional representation of process models by building tensors of activity-activity-constraint relations which can be fed into and learnt by convolutional long short-term memory networks (ConvLSTMs) \cite{DBLP:conf/nips/ShiCWYWW15}, an architecture capable of learning the interaction effects of said high dimensionality over time, to make predictions.
It is shown by an experimental evaluation on two real-life event logs that the convolutional LSTMs are indeed effective at predicting future process models, both for windows of fixed length (in terms of the number of events occurring in that window) and a fixed number of windows per trace (with a variable number of events occurring), over a varying window size.

The paper is structured as follows.
Section \ref{sec:moti} motivates the research concepts and provides an overview of related work.
In Section \ref{sec:prel}, an overview of the concepts used for building the networks is introduced followed by the actual methodology in Section \ref{sec:pam}.
In Section \ref{sec:eval}, PAM is verified on real-life event logs and analysed for its performance.
Section \ref{sec:conc} summarizes the findings and discusses points for future work.
\section{Related work and motivation}
\label{sec:moti}
In this Section, the previous efforts on predictive process mining are covered, followed by a motivation of the proposed approach within this context.

\subsection{Predictive process mining and monitoring}\label{sec:ppmlit}
Process prediction is an important topic for organisations.
Among others, it allows for better preparation in terms of resources and scheduling and it allows to keep track of KPIs that can be monitored by business constraints \cite{di2022predictive}.
In the process mining literature, various approaches have been used both for next-in-sequence prediction, as well as remaining time or time-to-next-activity prediction \cite{rama2021deep}.
A comprehensive overview of existing predictive process monitoring techniques has been devised in \cite{DBLP:journals/corr/TeinemaaDRM17,DBLP:journals/tsc/Marquez-Chamorro18,rama2021deep}. 
Besides these objectives, defining and monitoring other (control flow) objectives in processes has been considered as well.
\cite{DBLP:conf/caise/MaggiFDG14} and \cite{DBLP:journals/tsc/Francescomarino19} predict the outcome of business goals expressed in Linear Temporal Logic using decision trees, optionally supported by a trace clustering pre-processing step. 
E.g., goals could include $\Box(\text{accepting order}\to\Diamond\text{send invoice})$ to predict whether invoices have been adequately managed within an order process, or $\Diamond(\text{case is accepted})$ to predict whether a case is eventually accepted for payout in an insurance process.
This work has been further developed in \cite{DBLP:conf/bpm/Francescomarino17} in which prior knowledge regarding the process' future development is used to improve LSTM predictions. 
In \cite{DBLP:journals/eswa/ChamorroRCT17}, decision rules concerning mainly the data variables present in an event log are generated over multiple windows of an event log to make predictions regarding future values.
These rules are used to evaluate and predict future key performance indicators.
\cite{DBLP:journals/corr/TeinemaaDRM17} covers various aggregation and encoding mechanisms for traces as well as various classifiers which can be used for goal-oriented process prediction.

There exist many approaches which are capable of predicting these objectives.
One of the earlier works on predictive process mining introduced an approach based on abstractions to obtain finite state machines of a varying complexity which can be used to predict remaining time  \cite{DBLP:journals/is/AalstSS11}.
In a related approach, \cite{DBLP:journals/kais/LakshmananSDUK15} create Markov models to predict the next-in-sequence activity in various choice splits in process models.
\cite{DBLP:journals/misq/BreukerMDB16} use grammatical inference to create a probabilistic model on past behaviour to predict next-in-sequence activities.
The work of \cite{DBLP:journals/dss/EvermannRF17,DBLP:conf/ssci/NavarinVPS17} and \cite{DBLP:conf/caise/TaxVRD17} investigated the usefulness of LSTM neural network architectures for predicting next-in-sequence activities, and the remaining execution time based on recurrent neural networks. The introduction of neural networks caused a leap in predictive performance \cite{rama2021deep}.
It also enables left-in-trace prediction which goes beyond predicting just the next activity \cite{DBLP:conf/bpm/0001DR19}.
The effectiveness of deep neural networks was researched by \cite{DBLP:conf/wecwis/MehdiyevEF17} to predict the next-in-sequence activity using auto-encoder/decoder networks.
\cite{lin2019mm} introduces extra activity attribute information to improve prediction which allows not only to predict the next-in-sequence activity, but also its data attributes such as, e.g., the loan amount of a loan offer activity or the resource executing the activity, using encoder-decoder LSTMs.
\cite{DBLP:conf/icpm/Pasquadibisceglie19,pasquadibisceglie2020predictive} represent traces as images by converting them into a set of prefixes and employ convolutional neural networks to predict next-in-sequence activities.
An embedding approach for activities, traces, and process models has been proposed by \cite{DBLP:conf/bpm/KoninckBW18} based on random walks to discover the relationship of activities in a process.

In recent efforts, a leap towards process-wide predictive efforts have been made. 
\cite{park2020predicting} create system-wide transition systems to better model the interdependencies of cases on each other to improve remaining time prediction both at the trace and model level.
\cite{pasquadibisceglie2022promise} build a predictive algorithm which helps process discovery algorithms to select relevant information given the overwhelming amount of data generated by processes over time.
\cite{smedt2021process} introduce the concept of process model forecasting, which rather than predicting objectives or KPIs per case forecast directly-follows relations between activities to predict the future state of the process model (system).

Processes can be represented by a variety of models, including Petri nets \cite{murata1989petri} and BPMN \cite{BPMN}. However, these models are hard to track over time given they rely on a changing structure which is supported by XOR- and -AND-splits, as well as other concepts which make it hard to grasp them in a straightforward mathematical representation over time \cite{sommers2021process}. Declarative process models such as Declare models \cite{pesic2007declare}, on the other hand, rely on a fixed set of constraint templates which are easy to monitor (over time).
Various works on mining and monitoring declarative process models and constraints exist.
In \cite{DBLP:conf/otm/MaggiBCS13}, the on-line discovery of declarative models was adapted to detect changes in constraints due to concept drift and \cite{DBLP:conf/bpm/BurattinCM14} proposed an approach to treat processes as movies by tracking the violation status of declarative process constraints for event streams.
\cite{cecconi2022measuring} proposes a framework to track declarative constraints over an event log and score them using a wide range of measures.
Finally, \cite{yeshchenko2021visual} uses a map of declarative constraints over time to detect change points for drift detection.

\subsection{Processes-As-Movies}
On the one hand, it is hard for regular next-in-sequence-based approaches to look far ahead into a future case/trace as illustrated in \cite{DBLP:conf/bpm/0001DR19}.
This can be mitigated by using various solutions such as using more prefix information to detail the surroundings of the prediction \cite{DBLP:conf/bpm/0001DR19,DBLP:conf/icpm/Pasquadibisceglie19}.
On the other hand, it is hard to have the same high magnitude of predictive accuracy when multiple objectives over multiple cases are targeted at once, e.g., process model forecasting predicts many directly-follows relations at once \cite{smedt2021process}.

In this work, we propose an alternative hybrid approach of using model-based features per case/trace.
A historic process trace is divided into windows in which for each activity pair it is checked whether a set of declarative constraints hold (including unary constraints between an activity and itself).
This allows predictive models to learn both how constraints between activities are evolving over these windows within the trace, but also at an aggregate level.
This mimics not just creating an image, but a movie out of the process in the form of consecutive representations of a process model which can be learned by the overarching predictive model creating inferences over the full set of cases.
Note that in the extreme cases when a single activity is considered as a single window `standard' case-based predictive process monitoring is obtained, while very wide windows over long lengths of the trace provide long-term process model predictions/forecasts, hence covering a wide part of the predictive process monitoring-process model spectrum.

\begin{figure*}
	\centering
	\includegraphics[width=0.8\textwidth, trim={0 0cm 0 0cm}]{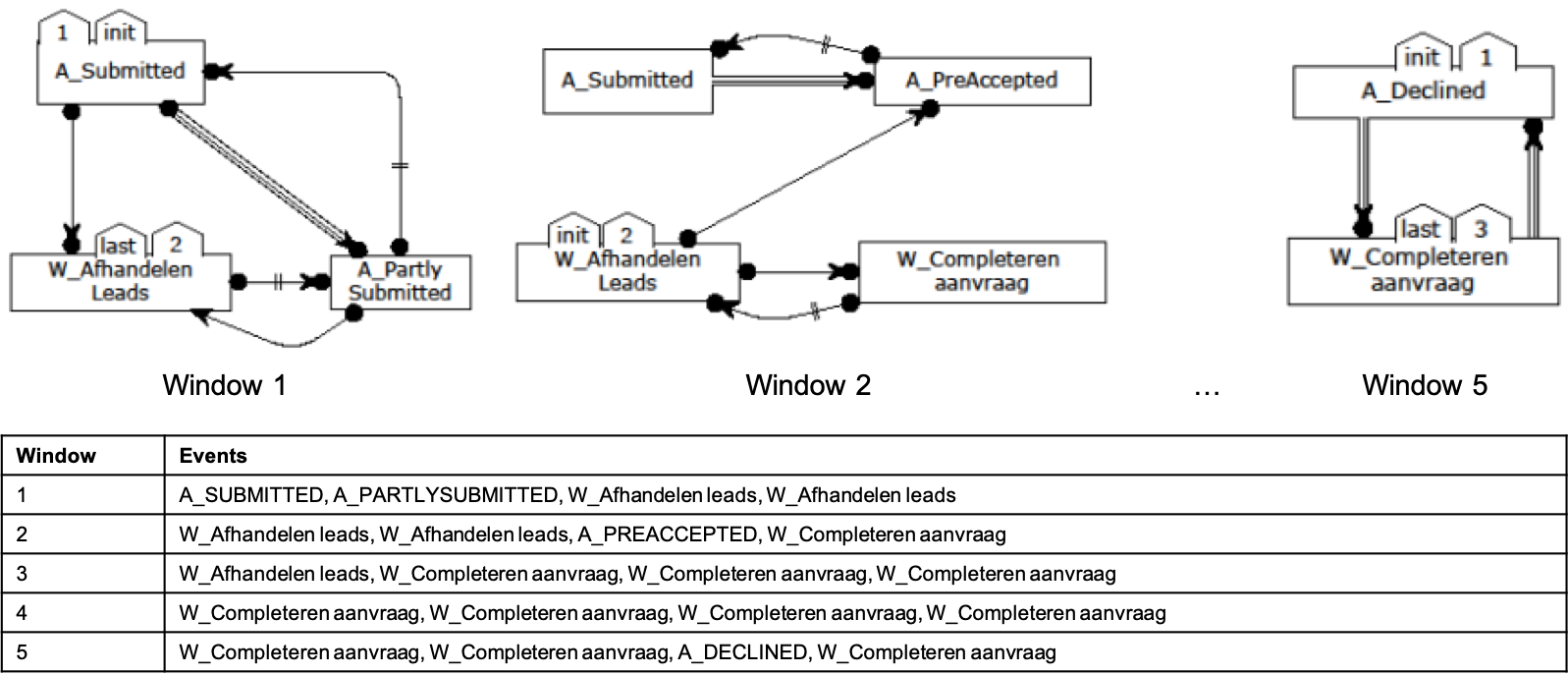}
	\caption{5 windows of length 4 in a trace of the 2012 BPI Challenge log and their corresponding process models.}
	\label{fig:example}
\end{figure*}

The setup allows to answer particular (control-flow) objectives (in LTL) similar to the ones outlined in Section \ref{sec:ppmlit}. 
Rather than using predefined objectives, however, in PAM a multi-activity result is provided in the form of relations between all activities at once.
Consider for example a loan application process which we base on the well-known example of the 2012 BPI Challenge\footnote{\url{https://data.4tu.nl/repository/uuid:3926db30-f712-4394-aebc-75976070e91f}}.
It is useful to know what activities are likely to occur next, e.g., whether an application will be rejected in the future (A\_Declined), however, this is a fine-granular prediction.
It is also useful to know what the underlying model is that generates the next activities.
For example, rather than knowing when and if a loan will be accepted or rejected, it can be revealing to know that there exists an underlying cause in the form of an activity that has not occurred, or the occurrence of a particular activity before the claim was decided on.
In Figure \ref{fig:example}, the loan application process of the 2012 and 2017\footnote{\url{https://data.4tu.nl/repository/uuid:5f3067df-f10b-45da-b98b-86ae4c7a310b}} BPI Challenge (which will be used in Section \ref{sec:eval}) is used as an illustration.
We can see that the final window, which is our prediction target, also contains information about when and how many times the activities are executed (\textit{A\_Declined} happens first in the window (init), and just once, \textit{W\_Completeren aanvraag} is executed 3 times and is the last in the window) and in what order (after each \textit{A\_Declined}, \textit{W\_Completeren aanvraag} has to happen next due to the \textit{chain response} constraint).
While previously proposed predictive process monitoring techniques build powerful models with activities as time steps and process model forecasting time intervals spanning hours, days, potentially weeks, PAM focuses on a granularity in between by dividing traces into smaller windows, in this example 5 windows, in which a small process model exists.
The underlying idea is to predict the next relations between activities that represent the model generating the behavior of the future of the trace.
This allows process owners to obtain predictions based on, e.g., whether an application was rejected because of the absence of a prior investigation (W\_Afhandelen Leads), the presence of a particular preceding check (A\_PreAccepted preceding A\_Submitted), or the inappropriate sequencing of events (e.g. an alternating sign-off cycle to adhere the four-eyes principle) all at the same time.
Furthermore, it allows to predict whether certain constraints which are monitored will be holding in the further execution of a process, e.g., the chain response constraint in window 5, offering a predictive alternative to \cite{DBLP:conf/bpm/BurattinCM14}.

By making use of the Declare language \cite{pesic2007declare}, it is possible to populate the windows with relations between activities, e.g., whether an activity will happen right after another, whether a particular activity will occur (a given number of times), and whether particular behaviour will not occur.
Other types of relations could be used as well, e.g., the 4C spectrum \cite{DBLP:conf/apn/PolyvyanyyWCRH14}, behavioural constraints \cite{weidlichPDMW11}, or incidence matrices \cite{weijters2006process}.
Nevertheless, the Declare base provides a well-rounded and comprehensive set of patterns that cover various unary and binary behavioural relationships which has seen a vast number of successful applications in mining and predicting processes.
The relations over the activity set create a vast feature space, which needs to be adequately modeled for prediction for which we explore various neural network-based architectures, most notably convolutional recurrent neural networks used in the analysis of moving images.
\section{Preliminaries}
\label{sec:prel}
In this section, the concept of traces, windows, and neural networks are introduced.

\subsection{Event logs, traces, and constraints}
Event logs are employed throughout process mining and analysis \cite{van2016data}.

\begin{definition}
A trace $t$ is a sequence of executions $t=\langle e_1, e_2,...,e_n\rangle$ of length $n$, where each execution in trace $t$ $e_{it}$ is of a particular event type from a set of activities $A$.
\end{definition}
I.e., $t \in A^*, \forall t \in\mathcal{L}$, where $A^*$ is the power set or language of all possible finite executions of activities.
\begin{definition}
An event log $\mathcal{L}$ is a list of traces.
\end{definition}
Consider Table \ref{tab:eventlog}, which displays an event log containing 5 traces.
\begin{table}
\centering
\begin{tabular}{|r|ll|}
	\hline
	\multicolumn{1}{|l|}{\textbf{ID}} & \textbf{Trace} & \textbf{windows} \bigstrut\\
	\hline
	\textbf{1} & abaabcdad & ab,aa,bc,dad \bigstrut[t]\\
	\textbf{2} & abaad & a,b,a,ad \\
	\textbf{3} & abaadc & a,b,a,adc \\
	\textbf{4} & cdaad & c,d,a,ad \\
	\textbf{5} & dabddd & d,a,b,ddd \bigstrut[b]\\
	\hline
\end{tabular}%
\caption{Example of an event log containing traces containing windows.}
\label{tab:eventlog}
\end{table}

\begin{definition}
A window is a subsequence of a trace denoted $t_w=\langle e_1, e_2,...,e_m \rangle \sqsubseteq t$ of subsequent executions, i.e, $\forall i, i+1\in[1,|t_w|],\, \forall j, j+1 \in [1,|t|], e_i=e_j\implies e_{i+1}=e_{j+1}$.
Hence, a trace $t$ can be divided into $n$ windows $t = \langle t_{w_1}, t_{w_2}, ...,t_{w_n}\rangle$ where $t_{w_i}\cap t_{w_j}=\emptyset,\, \forall i,j\in [1,n]$.
\end{definition}
In Table \ref{tab:eventlog}, the traces are divided in $n=4$ windows. 

\subsection{Neural network topologies}
Neural networks allow to model high-dimensional input and output spaces using an architecture of single or multiple layers of hidden neurons.
This architecture can be used for classification and/or regression, but also to create latent feature spaces using embeddings \cite{mikolov2013distributed}, or auto-encoder/decoders \cite{vincent2008extracting}.

Time-based and sequential data can be modelled using recurrent neural networks, which employ feedback loops between neurons to capture longitudinal dependencies.
Several extensions exist to account for the vanishing gradient problem such as long short-term memory networks (LSTMs)\cite{hochreiter1997long}.
An LSTM is a recurrent neural network with an intricate node structure to avoid the vanishing gradient problem \cite{DBLP:journals/neco/HochreiterS97}, meaning it can recover information over long distances. 
An LSTM unit takes a single vector at timestep $x_t$ as input and returns a hidden state $h_t$ based on the previous hidden state $h_{t-1}$ as follows:

\begin{align*}
f_t=\sigma(W_f[h_{t-1},x_t]+b_f),\\ i_t=\sigma(W_i [h_{t-1},x_t]+b_i),\\ \Tilde{C}=tanh(W_c[h_{t-1},x_t]+b_c), \\
C_t=f_t \odot C_{t-1}+i_t \odot \Tilde{C}_t,\\
o_t=\sigma(W_o [h_{t-1},x_t]+b_o),\\
h_t=o_t \odot tanh(C_t)
\end{align*}

with $\sigma(x)=\frac{1}{1+e^{-x}}$ the sigmoid function, $\odot$ the element-wise Hadamard product, and $W$ learnable weights and $b$ learnable biases.
$f_t$ indicates the forget gate to select which information should enter the current state, $i_t$ is the input gate controlling what information should be updated, and $o_t$ is the output gate which controls what information should be carried to the next state.
For the outcome of a full network of $n$ LSTM layers of $p$ units, e.g., for prefix $\pi$ of length $n$, we write $LSTM(\pi)=\langle h_1,\dots,h_n\rangle\in\mathbb{R}^{n\times p}$ to denote the sequence of outputs. 

Furthermore, the concept of convolution is used to alleviate the complexity inherent to high dimensionality by using filters that focus on particular parts of the feature space, which has been used successfully in areas such as image recognition and modelling \cite{badrinarayanan2017segnet}.
They are especially gaining traction in the field of computer vision because of their capabilities to capture multi-dimensional data, often needed for moving images in different colour bands have made them the best-performing choice for image recognition, classification, and prediction \cite{DBLP:journals/cacm/KrizhevskySH17}.
Convolutions can be used to make lower-dimensional representations of 1D vectors (e.g. sensor information such as electrocardiograms \cite{DBLP:journals/corr/abs-1905-03554}), 2D matrices (e.g. images \cite{lecun1995convolutional}) or 3D matrices (e.g. moving pictures or videos \cite{tran2018closer}). Convolutions are constructed through the filtering of tensors by striding through the tensor in a lower dimensionality. E.g., for a 2D tensor of $25\times25$, kernels of $K=5$ can be used to construct $5\times5$ filters. After creating kernels covering the whole tensor they are pooled, e.g., through maximum or average pooling. Convolutions allow to obtain lower dimensionality while retaining lateral relations such as temporal and spatial correlations in images and video. 

Various architectures can be combined as well.
Encoder-decoder LSTMs are used for machine translation \cite{DBLP:conf/emnlp/ChoMGBBSB14}, and network evolution modelling \cite{chen2019lstm}.
They map high-dimensional (typically word) vectors to output sentences, similar to the presence of constraints in tensors as will be discussed in Section \ref{sec:pam}.
They create sequence embeddings by, layer per encoding/decoding layer, capturing latent dimensions in the input/output time steps. The power of this network topology is based on the number of latent dimensions that are used before feeding the encoded data to the LSTM core layer, and how many layers are used to reduce the dimensionality of the data.
A combination of encoder-decoder convolutions is possible as well \cite{badrinarayanan2017segnet} where subsequent convolutions are getting smaller and smaller in the encoder part and increase in size in the decoder part.
Finally, the combination of LSTMs and convolutions, convolutional LSTMs (ConvLSTMs), allows for the joint optimisation of the architectures which has been successfully used for precipitation prediction and video analysis \cite{DBLP:conf/nips/ShiCWYWW15,DBLP:conf/nips/FinnGL16,DBLP:conf/cvpr/LiangH15}.
This topology is based on a sequential input that goes through various convolutional recurrent layers which allows for the dynamic analysis of high-dimensional data in every layer of the network and not just in the input and/or output.
This avoids that the input and output layers need to be fully connected over the whole feature space, and results in a more compact architecture.
Similar to other convolutional neural networks, max pooling with filters is used to capture smaller parts that contain a particular correlation in the high dimensional input and relate them to each other throughout the network.

\section{Methodology and Implementation}
\label{sec:pam}
The main PAM methodology consists of two main steps.
First, the traces in the event log are featurized to be inputted into the inference mechanism.
Secondly, the resulting transformed event logs can be used as input tensors to neural networks.

\subsection{Feature Generation}
The goal of the approach is to capture relationships between activities over time. To this purpose, we mine for the (binary) presence of a relationship between activities $\mathbf{R} \in \{0,1\}^{A\times A\times C}$, where $C$ is a set of relation types.
$\mathbf{R}_t$ denotes the matrix that exists between activities for trace $t$.
To introduce the dynamic aspect, we divide traces into windows, for which the relations can be denoted $\mathbf{R}_{t_{w}}$ for all traces in $\mathcal{L}$ at various windows.
Hence, for every window $w\in t$ in every trace $t\in\mathcal{L}$, we can obtain a binary vector of length $|C|$ for every pair of activities denoting the presence of constraint $c\in C$. 
In the following sections, we discuss what type of relations are suitable, and how traces are divided into windows.

\subsubsection{Activity relations}
To capture the evolution of a process, an appropriate featurization step is needed which can represent its underlying changes.
In process mining, both procedural and declarative languages are used.
Especially Petri nets \cite{murata1989petri} are successfully employed by various well-known algorithms such as Alpha Miner \cite{van2004workflow}, or Inductive Miner \cite{leemans2013discovering}.
However, the relationships between activities in a Petri net cannot be captured straightforwardly. 
Due to the use of places and hidden transitions to incorporate long-distance dependencies and concurrency, there is no easy mapping possible from model constructs to a fixed set of variables as activities might be connected through many places or even places and hidden activity pairs \cite{sommers2021process}.
The same holds for other procedural models such as causal nets \cite{van2011causal}, or BPMN \cite{BPMN} given they have similar (control flow) semantics.
Other mining algorithms such as Heuristics Miner \cite{weijters2006process} which rely on dependency graphs or directly-follows graphs, do have these activity-to-activity relationships similar to declarative process models such as Declare \cite{pesic2007declare} or DCR Graphs \cite{hildebrandt2011declarative}, however, they lack the expressiveness of the latter languages.
Given the widespread support and the ability to efficiently mine Declare constraints using \cite{de2019mining}, we opt to use them to capture the relations between our activities.
An overview of the constraints expressed in LTL and regular expressions used can be found in Table \ref{tab:declare_formal}.
\begin{table}[tb]
\centering
\scriptsize
	\resizebox{0.9\textwidth}{!}{
		\begin{tabular}{|p{3cm}|p{3.5cm}|p{5cm}|}
			\hline
			\textbf{Template} & \textbf{LTL Formula \cite{pesic2008constraint}} & \textbf{Regular Expression \cite{westergaardunconstrainedminer}}  \tabularnewline
			\hline\hline
			Existence(A,n) & $\Diamond(A\wedge\bigcirc(existence(n-1,A)))$ & .{*}(A.{*})\{n\} \tabularnewline\hline
			Absence(A,n) & $\neg existence(n,A)$ & {[}\^{ }A{]}{*}(A?{[}\^{ }A{]}{*})\{n-1\}  \tabularnewline\hline
			Exactly(A,n) & $existence(n,A) \wedge absence(n+1,A)$ & {[}\^{ }A{]}{*}(A{[}\^{ }A{]}{*})\{n\} \tabularnewline\hline
			Init(A) & $A$ & (A.{*})? \tabularnewline\hline
			Last(A)  & $\Box(A\implies\neg X\neg A)$ & .{*}A \tabularnewline\hline
			Responded existence(A,B) & $\Diamond A\implies\Diamond B$ & {[}\^{ }A{]}{*}((A.{*}B.{*}) $|$(B.{*}A.{*}))?\tabularnewline\hline
			Co-existence(A,B) & $\Diamond A\impliedby\Diamond B$ & {[}\^{ }AB{]}{*}((A.{*}B.{*}) $|$(B.{*}A.{*}))?\tabularnewline\hline
			Response(A,B) & $\Box(A\implies\Diamond B)$ & {[}\^{ }A{]}{*}(A.{*}B){*}{[}\^{ }A{]}{*}   \tabularnewline\hline
			Precedence(A,B) & $(\neg B\, UA)\vee\Box(\neg B)$ & {[}\^{ }B{]}{*}(A.{*}B){*}{[}\^{ }B{]}{*} \tabularnewline\hline
			Succession(A,B) & $response(A,B)\wedge precedence(A,B)$ & {[}\^{ }AB{]}{*}(A.{*}B){*}{[}\^{ }AB{]}{*} \tabularnewline\hline
			Alternate response(A,B) & $\Box(A\implies\bigcirc(\neg A\, U\, B))$ & {[}\^{ }A{]}{*}(A{[}\^{ }A{]}{*}B{[}\^{ }A{]}{*}){*} \tabularnewline\hline
			Alternate precedence(A,B) & $precedence(A,B)\wedge\Box(B\implies\bigcirc(precedence(A,B))$ & {[}\^{ }B{]}{*}(A{[}\^{ }B{]}{*}B{[}\^{ }B{]}{*}){*}  \tabularnewline\hline
			Alternate succession(A,B) & $altresponse(A,B)\wedge precedence(A,B)$ & {[}\^{ }AB{]}{*}(A{[}\^{ }AB{]}{*}B{[}\^{ }AB{]}{*}){*} \tabularnewline\hline
			Chain response(A,B) & $\Box(A\implies\bigcirc B)$ & {[}\^{ }A{]}{*}(AB{[}\^{ }A{]}{*}){*} \tabularnewline\hline
			Chain precedence(A,B) & $\Box(\bigcirc B\implies A)$ & {[}\^{ }B{]}{*}(AB{[}\^{ }B{]}{*}){*} \tabularnewline\hline
			Chain succession(A,B) & $\Box(A\iff\bigcirc B)$ & {[}\^{ }AB{]}{*}(AB{[}\^{ }AB{]}{*}){*} \tabularnewline\hline
			Not co-existence(A,B) & $\neg(\Diamond A\wedge\Diamond B)$ & {[}\^{ }AB{]}{*}((A{[}\^{ }B{]}{*}) $|$(B{[}\^{ }A{]}{*}))?  \tabularnewline\hline
			Not succession(A,B) & $\Box(A\implies\neg(\Diamond B))$ & {[}\^{ }A{]}{*}(A{[}\^{ }B{]}{*}){*}  \tabularnewline\hline
			Not chain succession(A,B) & $\Box(A\implies\neg(\bigcirc B))$ & {[}\^{ }A{]}{*}(A+{[}\^{ }AB{]}{[}\^{ }A{]}{*}){*}A{*}  \tabularnewline\hline
			Choice(A,B) & $\Diamond A\vee\Diamond B$ & .{*}{[}AB{]}.{*} \tabularnewline\hline
			Exclusive choice(A,B) & $(\Diamond A\lor\Diamond B)\wedge\neg(\Diamond A\land\Diamond B)$ & ([\^{ }B]*A[\^{ }B]*) $|$.{*}{[}AB{]}.{*}({[}\^{ }A{]}{*}B{[}\^{ }A{]}{*}) \tabularnewline
			\hline
		\end{tabular}
	}
	\caption{An overview of Declare constraint templates with their corresponding LTL formula and regular expression.}
	\label{tab:declare_formal}
\end{table}
The benefit of using Declare constraints is that they cover both unary and binary relations, meaning they can also cover the behaviour of a single activity.
Furthermore, the constraints' automata can be converted into Petri nets if desired \cite{prescherdeclarative,desmedt2015ri}.

\subsubsection{Windows}
A crucial part of obtaining enough information to make predictions towards future process execution relies on the amount of information that is available, and especially how it is structured when fed to predictive models.
To this purpose, various approaches have been introduced for trace bucketing, sequence and prefix extraction and encoding, which have been summarised in \cite{DBLP:journals/corr/TeinemaaDRM17}.
The prefix extraction typically relies on iteratively introducing more information about the prefix as the execution develops in the system, making the prefixes grow in length.
Often, traces are grouped to ensure models do not have to deal with vast discrepancies in trace length, or the types of activities present in the trace.
For PAM, we use a window-based approach to capture the behaviour of a process in various stages of its development.
To this purpose we approach a trace in two ways; either a fixed window size is used, or a fixed number of windows.
These serve different purposes.

A fixed window size allows to collect a number of events and base a future prediction on the previous windows of the same size.
If a predictor was created for, e.g., windows of length 5, one could predict as soon as 5 events have happened to predict the process behaviour in the next window of 5 events.
This entails, however, that to deal with different number of windows, e.g., in case we have already collected 10 events to predict the following window of 5 events, we might have to apply different padding to make the input steps of the LSTMs of equal length.
To this purpose, we will also investigate the use of different models for a different number of windows of the same size, similar to \cite{DBLP:conf/bpm/LeontjevaCFDM15}.
A fixed number of windows on the other hand allows to divide a trace regardless of whether it has attained a sufficient length.
Given that there is a variety of trace lengths available for training, and given that they can be used with a pre-trained model for a particular number of windows, PAM should be able to deal with the discrepancies that exist between trace length.
For example, using a model trained on traces with a fixed number of windows of 5 allows to use both a trace of length 4 to predict the fifth execution step, and a trace of length 20 to predict the model in the fifth execution window.
Both window-creation approaches will be experimented with in Section \ref{sec:eval}, however, the latter approach is more flexible in terms of requirements on the trace.

Other techniques such as stages in processes \cite{DBLP:conf/caise/NguyenDHRM16} can be used to split traces as well, however, the stages at process level might not correspond directly with stages in a trace.
However, in the case of a fixed number of windows the use of equal input steps allows for a fair comparison across various trace lengths as process logs often contain long and short traces with different activities, and is easy to apply to any trace where $|t|\geq w$.
Besides, it does not spill any information into the training process.
In large event logs, shorter and longer traces will provide enough evidence of different types of execution patterns to make sure that windows in traces of different lengths are representative and learned by the recurrent neural network.

\subsubsection{Inference}
In order to do extract the Declare constraints efficiently, a window-based version of the interesting Behavioural Constraint Miner (iBCM) \cite{de2019mining} is used.
This technique mines for Declare constraints in an efficient manner by making use of basic string operations.
Compared to other declarative process mining algorithms such as Declare Miner \cite{maggi2011user} or minerFUL++ \cite{di2013two}, this approach does not focus on finding a  model by making use of support and confidence over all traces, but rather elicits the constraints per trace. 
This result would be roughly similar to using the former two approaches per trace with a very low support and with 100\% confidence, as only activities present in the traces are considered.
Also, iBCM is capable of retrieving the constraints over a predefined number of windows.
Traces are divided in a number of windows $w$ where $|t_{w_{i}}|=\lceil\frac{|t|}{w}\rceil$. The last window's size varies according to the discrepancies between the window and trace length.
E.g., $t=\langle a, b, c, d, e\rangle$ for $w=3$ would be split up in the following windows: $t_{w_1}=\langle a,b\rangle, t_{w_2}=\langle c, d\rangle, t_{w_3}=\langle e\rangle$.
Note that this is a coarse way of using the window principle and it would be interesting to pursue a more tailored split, however, in case a large number of training points are available the difference in window sizes will be learnt by the model to overcome this issue.

Since constraints are mined for single traces, the models that can be composed from generating the product of the separate constraint's automata, are always calculable \cite{DBLP:conf/bpm/CiccioMMM15}. 
I.e., constraints found in a single trace are never conflicting.
Hence, the inference step will not have evidence of inconsistent models and will be capable of producing sound output models if the resulting neural network is capable of reproducing the original input data (completely).

Not every constraint listed in Table \ref{tab:declare_formal} is as suitable for the envisioned application.
First of all, \textit{not chain succession} is a constraint which is satisfied for a high number of activity pairs quickly, as it is a negative constraint for which counter evidence is scarce.
In terms of unary constraints, only \textit{absence(a,1)}, \textit{exactly(a,1/2)}, and \textit{existence(a,3)} are used to limit the size of the feature vectors.
For the same reason, \textit{succession} and its two other variants are not used because they overlap with other constraints completely, which renders them redundant (\textit{response} and \textit{precedence}).
Given their similarity (on a single trace), only \textit{exclusive choice} is used and not \textit{not co-existence} for the latter would be overlapping with \textit{absence} constraints.
Note that unary constraints are included as a relationship between the activity and itself to allow the inclusion in the matrix structure.
The inclusion of the negative constraints \textit{absence} and \textit{not succession} are providing a process construct which is not available in other sequence generation based approaches using language modelling for next-in-sequence prediction.
Although LSTMs might train the absence of certain events in a particular sequence, PAM is explicit about this behaviour and provides more insight into the prediction compared to the former because of the inclusion of the constraint types.

The final result, i.e., the binary vectors denoting which constraints per activity pair per window are present, is very sparse as a vast number of possible relations between constraints and activity pairs exist. 
This makes the modelling challenging, however, neural networks are suitable to work with such a sparse high-dimensional input. 

\subsection{Predictive Network}\label{sec:4network}

Once the presence of all constraints is mined and stored per activity pair per trace, i.e., 3D matrices or tensors, a predictive model can be trained over the data.
To do this, we use two LSTM-based topologies capable of capturing high-dimensional inputs, an encoder-decoder setup with an increasing amount of layers, and a ConvLSTM with convolutions per time step.

The encoder-decoder setup is used by stacking layers of LSTMs with a decreasing and subsequent increasing number of neurons.
To this purpose the 3D activity-activity-constraint representation $R_{t_w}$ is transformed into a flat one $R^*_{t_w}\in \{0,1\}^{A\cdot A \cdot C}$.
Then, a number of hidden encoder LSTM layers is introduced with a decreasing amount of neurons until a vector $z\in\mathbb{R}^e$ is obtained as an $e$-dimensional representation of the initial input vector.
Finally, $z$ is upscaled through the same number of decoder layers with an increasing amount of neurons, effectively mirroring the input stage towards the output.

\begin{figure}
	\centering
	\includegraphics[width=\textwidth, trim={10 10 10 10}]{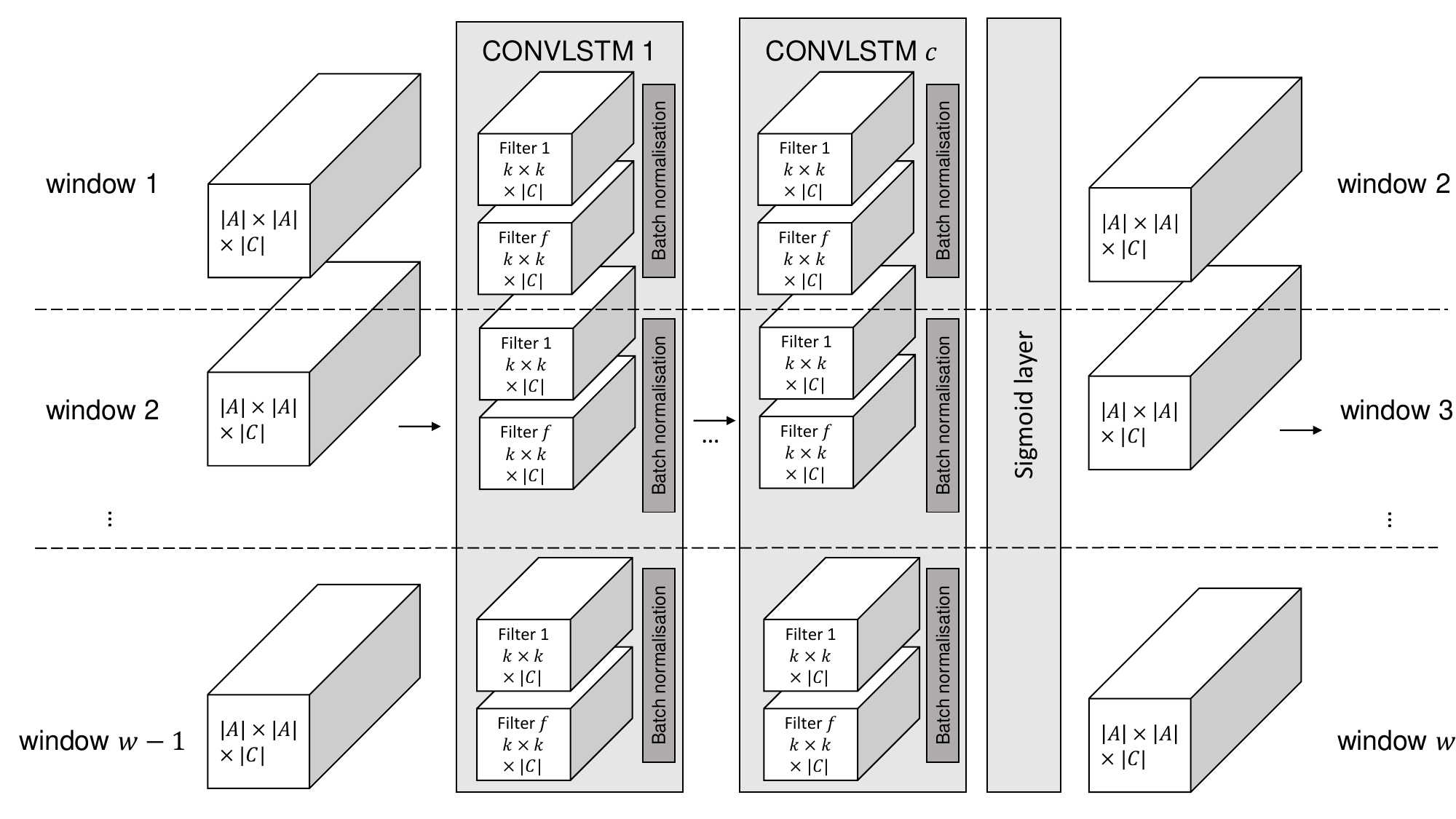}
	\caption{Overview of the main architecture of convolutional recurrent neural networks.}
	\label{fig:arch}
\end{figure}

The ConvLSTM setup is adapted to the case of $\mathbf{R}_{t_{w}}$ which is a 3D tensor which changes over time.
Time-based 3D convolutions are used as convolutions in the originally defined 2D setup of ConvLSTMs \cite{DBLP:conf/nips/ShiCWYWW15}:
\begin{align*}
f_t=\sigma(W_f * [\mathcal{H}_{t-1},\mathcal{X}_t] + W_{cf} \odot \mathcal{C}_{t-1} +b_f),\\ i_t=\sigma(W_i  * [\mathcal{H}_{t-1},\mathcal{X}_t] + W_{ci} \odot \mathcal{C}_{t-1} +b_i),\\ \mathcal{\Tilde{C}}=tanh(W_c * [\mathcal{H}_{t-1},\mathcal{X}_t]+b_c), \\
\mathcal{C}_t=f_t \odot \mathcal{C}_{t-1}+i_t * \mathcal{\Tilde{C}}_t,\\
o_t=\sigma(W_o * [\mathcal{h}_{t-1},\mathcal{X}_t] + W_{co}\odot \mathcal{C}_t +b_o),\\
\mathcal{H}_t=o_t \odot tanh(\mathcal{C}_t)
\end{align*}
with $*$ the convolution operator and the input $i_t$ a single $\mathbf{R}_{t_{w}}$.
There are a number of parameters to consider.
First of all, the input features are split up into out-takes of a lower dimension by using max pooling.
This results in a subset of activity pairs over all constraints.
The size of the filters, as well as how many filters are used to influence the granularity of the outcome.
To make the analogy with frames in a video, smaller frames are learnt to find, e.g., particular objects which deeper in the network can be combined to learn a particular larger object.
In processes, this is equivalent to learning a small set of activities, e.g., a block structure, which might later serve in a particular relationship with other blocks.
Larger filters are capable of capturing concepts moving over time faster, while smaller ones are better capable of processing slower evolving information.
Given the significant proportion of recurring constraints for a higher number of windows as will be reported in Section \ref{sec:eval}, we expect smaller kernels to work better on less-changing, many windows, and larger kernels to capture more information when fewer, more different windows are present as they will appear to be changing faster.
Secondly, the number of stacked ConvLSTM layers can be varied to obtain a deeper architecture.
The latter can be dedicated to smaller parts of the input and can have lower dimensionality.
This increases the expressiveness of the network and subsequently can increase the (predictive) performance of the network; however, it can also lead to overfitting.
Furthermore, it increases the number of parameters that need to be learnt, resulting in higher computation times.
After each ConvLSTM layer, batch normalisation is applied to boost overall performance \cite{DBLP:conf/icml/IoffeS15}.
The main architecture (already applied to the representation discussed in Section \ref{sec:pam}) is shown in Figure \ref{fig:arch}.
For a more detailed description of the architecture, we refer to \cite{DBLP:conf/cvpr/LiangH15} and \cite{DBLP:conf/nips/ShiCWYWW15}.

In the evaluation setup of the experiment in Section \ref{sec:eval}, flat encoder-decoders represent the most simple form of encoding such high-dimensional input vectors and are compared with ConvLSTMs which are capable of extracting the correlations between constraints and activities over time as they are (better) preserved through the convolutions. The goal is to evaluate whether the extra computational power required for these convolutions justifies any gains in predictive performance.

The final layer of each network exists of a 3D sigmoid layer which maps the results to a $|A|\times|A|\times|C|$ output probability which can serve as a binary prediction after applying a threshold.
Also, the network's optimisation parameters need to be chosen, i.e., the loss function, the optimizer, and the number of epochs, as they have a strong impact on the results of a (recurrent) neural network.
Note that the number and size of windows affect the performance of the technique significantly as well.
In the extremest case, the window sizes can be set equal to the number of events in a trace, which makes it possible to create a standard LSTM similar to the setup of \cite{DBLP:conf/caise/TaxVRD17}, as only \textit{exactly} and \textit{absence} constraints will be found in every window. 
\section{Evaluation}
\label{sec:eval}
In this section the setup and results of the experimental evaluation are discussed.
In addition, the interpretation and impact of constraint types are analysed.
The main goal is to test whether PAM is capable of learning traces as sequences of windows. 
More specifically, we investigate whether PAM is capable of predicting the constraints that will be present in subsequent windows accurately.
As a benchmark, we compare the approach with LSTMs for remaining trace prediction as used in PPM.

\subsection{Setup}
\subsubsection{Network topologies and implementation}
PAM has been implemented as a feature generation technique in Java, and a deep learning network architecture in the Keras Python library\footnote{\url{https://keras.io/}}.
The implementations can be found online\footnote{\url{https://github.com/JohannesDeSmedt/processes-as-movies}}.

As indicated before, two types of recurrent neural network topologies are used.
Firstly, encoder/decoder networks were used to work with the high-dimensional input. Two hyperparameters were used: the dimensionality of the input layer, and the number of encoding/decoding layers.
The former was set to powers of 2, i.e., 64, 128, 256, and 512. 
Besides, 1-4 layers of encoding/decoding were used, where every subsequent layer is half of the size of the former.
Secondly, ConvLSTMs were used with three hyperparameters; kernel size and filter size are convolutional neural network parameters, set at 4, 8, and 12, and 1-4 CONVLSTM layers were used.
Note that further hyper-parameter optimisation could still improve the results, as this is an essential part of the learning process \cite{DBLP:journals/is/Francescomarino18}.
The experiments were run both for a fixed number of windows, and a fixed window length, both set at 2, 5, 10 to obtain insight in the effect of varying window sizes and lengths.
Note that a fixed number of windows of 2 splits a trace in half and predicts the presence of constraints in the second half of the trace, which is a hard task for an LSTM as there is no potential to propagate a lot of past information.  
The number of epochs was set at 10 for the fixed number of windows dataset and 20 for the shorter logs used for the fixed window size as the results did not change significantly when using higher values (tested at 40 and 60) for either approach and to make a trade-off in terms of results/performance.
Given that the input dimensionality is high (e.g. for BPI 17 with $|A|=26$ we have time steps of dimensionality $26\times26\times14=9,464$), batch sizes needed to be sufficiently small (10-20 traces) to fit in memory.

Two optimizers were used, Nadam \cite{ruder2016overview} and ADADELTA \cite{DBLP:journals/corr/abs-1212-5701}, both with the binary cross entropy measure as a loss function.
In all cases, Nadam outperformed ADADELTA and is reported in the results.
Activity and kernel regularisation \cite{zaremba2014recurrent} were applied in various layers as well as the final layer, but yielded no better results and are also not reported.

All models were run on a single NVidia GeForce GTX1070 Ti with 8GB of video memory and 2,432 CUDA cores. 
All calculations can be performed on a standard desktop computing setup within reasonable time. The timings reported are in seconds for 1 epoch.

\subsubsection{Data}
Two popular publicly available event logs are used, i.e., the 2012 and 2017 BPI Challenge logs.
As illustrated in Section \ref{sec:moti}, they handle a loan application process, which consists of opening an application, handling it, and finally making a decision on its status.
The results of applying iBCM, as well as statistics on the number of activities and traces can be found in Tables \ref{tab:dec2vecfnw} and \ref{tab:dec2vecfws}. 
The performance of iBCM in terms of generating the features on the event logs used in the evaluation section for a fixed number of windows, and a fixed window size respectively.
The results were obtained with a Java 8 Virtual Machine on an Intel Xeon E3-1230 (v5) CPU with 32GB DDR4 memory.
Overall, the technique is capable of quickly generating a vast amount of constraints present in various sizes of windows within a trace which can serve as input to the later inference stage.
To get an idea of how similar windows are, the overlap of recurring constraints between subsequent windows, i.e., windows 1 and 2, 2 and 3, and so on, is listed as well.
For the fixed number of windows version, no extra pre-processing needs to be taken as every trace that has at least as many elements as the number of windows will be used.
This means that some traces might be too short for, e.g., higher window sizes.
This technique results in both very small and very large windows to be generated.
For the fixed window size version, traces are mined for a number of windows present to avoid any padding in the inference stage later.
The event logs were divided into subsets where 5 windows are present for window size 2, 2 windows in case of window size 5, and 1 window in case of window size 10.
Only event logs with at least 2 windows (at least 2 windows are needed to have at least 2 time steps to train the LSTMs) were considered.
Hence, only traces of at least length 11 are used (to have at least 1 window of 10, and 1 of 1 event).

There is a significant difference between the event logs in terms of generated constraints, which is mainly due to the number of traces.
In Table \ref{tab:dec2vecfnw}, we see that the overlap is higher when more windows are used to divide a trace.
\begin{table}[tb]
\centering
\scriptsize
	\resizebox{\textwidth}{!}{
		\begin{tabular}{|l|r|r|r|r|r|r|}
			\hline
			\textbf{dataset} & \multicolumn{1}{l|}{\textbf{\#traces (min, max, avg.)}} & \multicolumn{1}{l|}{\textbf{\#windows}} & \multicolumn{1}{l|}{\textbf{ \#constraints }} & \multicolumn{1}{l|}{\textbf{ \#traces too short }} & \multicolumn{1}{l|}{\textbf{overlap}} & \multicolumn{1}{l|}{\textbf{time (s)}} \bigstrut\\
			\hline
			\multicolumn{1}{|r|}{} &       & 2     &        2,701,992  &                               -    & 0.447 & $<$1 \bigstrut[t]\\
			\textbf{BPI 2012} & \multicolumn{1}{l|}{13,087 (3, 175, 20)} & 5     &        2,486,721  &                         3,429  & 0.613 & 1 \\
			\textbf{($|A|=24$)} &       & 10    &        2,655,590  &                         6,106  & 0.705 & 1 \bigstrut[b]\\
			\hline
			\multicolumn{1}{|r|}{} &       & 2     &      13,546,340  &                               -    & 0.13  & 7 \bigstrut[t]\\
			\textbf{BPI 2017} & \multicolumn{1}{l|}{31,509 (10,180, 38)} & 5     &      11,228,795  &                               -    & 0.389 & 6 \\
			\textbf{($|A|=26$)} &       & 10    &      13,457,629  &                               -    & 0.619 & 10 \bigstrut[b]\\
			\hline
		\end{tabular}%
	}
	\caption{An overview of the performance and output of iBCM on the event logs used for evaluation for a fixed number of windows per trace.}
	\label{tab:dec2vecfnw}
\end{table}
Hence, the intuition holds that models tend to be more similar the closer they are in time.
The shorter traces of BPI 2012 also have a higher overlap than the traces in the BPI 2017 event log.
It will be interesting to see how this affects the results of the predictive models below, i.e., whether higher overlap leads to higher levels of accuracy/precision.
In Table \ref{tab:dec2vecfws}, we see a similar effect: deriving (more) windows of length two results in higher overlap.
\begin{table}[tb]
\centering
	\resizebox{0.7\textwidth}{!}{

\begin{tabular}{|c|c|r|r|r|r|r|}
	\hline
	\multicolumn{1}{|p{3.68em}|}{\textbf{dataset}} & \multicolumn{1}{l|}{\textbf{window length}} & \multicolumn{1}{l|}{\textbf{\#windows}} & \multicolumn{1}{l|}{\textbf{\#traces}} & \multicolumn{1}{l|}{\textbf{ \#constraints }} & \multicolumn{1}{l|}{\textbf{overlap}} & \multicolumn{1}{l|}{\textbf{time (s)}} \bigstrut\\
	\hline
	\multicolumn{1}{|c|}{\multirow{15}[6]{*}{\textbf{BPI 2012}}} & \multirow{5}[2]{*}{2} & 6--10 & 1,351 &           263,584  & 0.618 & $<$1 \bigstrut[t]\\
	&       & 11--15 & 1,847 &           750,706  & 0.644 & 1 \\
	&       & 16--20 & 1,525 &           826,634  & 0.662 & 1 \\
	&       & 21--25 & 883   &           589,873  & 0.652 & $<$1 \\
	&       & 26--30 & 458   &           385,314  & 0.665 & $<$1 \bigstrut[b]\\
	\cline{2-7}      & \multirow{5}[2]{*}{5} & 3--4  & 1,351 &           486,917  & 0.351 & $<$1 \bigstrut[t]\\
	&       & 5--6  & 1,847 &           515,539  & 0.521 & $<$1 \\
	&       & 7--8  & 1,525 &           537,837  & 0.545 & $<$1 \\
	&       & 9--10 & 883   &           377,288  & 0.513 & $<$1 \\
	&       & 11--12 & 458   &           237,295  & 0.517 & $<$1 \bigstrut[b]\\
	\cline{2-7}      & \multirow{5}[2]{*}{10} & 2     & 1,351 &           185,897  & 0.048 & $<$1 \bigstrut[t]\\
	&       & 3     & 1,847 &           516,404  & 0.084 & $<$1 \\
	&       & 4     & 1,525 &           495,542  & 0.145 & $<$1 \\
	&       & 5     & 883   &           337,116  & 0.198 & $<$1 \\
	&       & 6     & 458   &           202,793  & 0.267 & $<$1 \bigstrut[b]\\
	\hline
	\multicolumn{1}{|c|}{\multirow{15}[6]{*}{\textbf{BPI 2017}}} & \multirow{5}[2]{*}{2} & 6--10 & 2,278 &           623,853  & 0.679 & $<$1 \bigstrut[t]\\
	&       & 11--15 & 10,417 &        4,315,163  & 0.79  & 4 \\
	&       & 16--20 & 7,274 &        4,155,367  & 0.776 & 6 \\
	&       & 21--25 & 5,388 &        3,878,768  & 0.775 & 6 \\
	&       & 26--30 &    3,085  &        2,772,172  & 0.785 & 2 \bigstrut[b]\\
	\cline{2-7}      & \multirow{5}[2]{*}{5} & 3--4  & 2,278 &           486,917  & 0.267 & $<$1 \bigstrut[t]\\
	&       & 5--6  & 10,417 &        3,041,851  & 0.43  & $<$1 \\
	&       & 7--8  & 7,274 &        2,703,384  & 0.446 & $<$1 \\
	&       & 9--10 & 5,388 &        2,403,113  & 0.448 & 1 \\
	&       & 11--12 &    3,085  &        1,639,413  & 0.465 & $<$1 \bigstrut[b]\\
	\cline{2-7}      & \multirow{5}[2]{*}{10} & 2     & 2,278 &           535,864  & 0.02  & $<$1 \bigstrut[t]\\
	&       & 3     & 10,417 &        3,066,199  & 0.063 & $<$1 \\
	&       & 4     & 7,274 &        2,588,962  & 0.101 & $<$1 \\
	&       & 5     & 5,388 &        2,218,077  & 0.157 & $<$1 \\
	&       & 6     &    3,085  &        1,445,214  & 0.226 & $<$1 \bigstrut[b]\\
	\hline
\end{tabular}
	}
	\caption{An overview of the performance and output of iBCM on the event logs used for evaluation for a fixed window size.}
	\label{tab:dec2vecfws}
\end{table}
The longer the traces and the more windows, the higher the overlap as well, however, this effect levels off quickly except for long windows (size 10).
Again, it will be interesting whether this will affect the modelling step.

\subsubsection{Evaluation criteria}
All experiments were performed using a 80\%/20\% training/test setup with another 20\% validation set during the training epochs.
To evaluate the neural network models, standard binary classification metrics are used to evaluate whether a constraint is predicted to be present correctly or not.
The area under the precision-recall curve (AP), the F(1)-score at the best threshold on the precision-recall curve, and the Area Under Receiver operating characteristics curve (AUC) are calculated to give a wide overview of whether the network is capable of predicting constraints holding or not holding in a window (true positives and negatives) compared to falsely predicting the presence or absence of constraints (false positives and negatives).
Given the sparsity of the matrices, i.e., only a few constraints hold between activity pairs throughout a particular time window, the accuracy can quickly gravitate towards very high numbers, affecting even the AUC. 
Therefore, the AP will be more revealing in terms of the power of the approach towards predicting the presence of constraints correctly (all the positive observations).

\subsection{Results}
In this Section, we first compare with a baseline LSTM approach. 
Next, we discuss the results stemming from both neural network approaches, and the effect of their parameterisation.

\subsubsection{Baseline LSTM approach}
In Table \ref{tab:lstms}, the results of using LSTMs for remaining trace prediction through the use of hallucination as proposed by \cite{DBLP:conf/bpm/0001DR19}.
For the predicted traces it is checked whether the constraints present in this predicted sequence of activities matches the actual constraints present in the actual sequence.
While not a direct comparison with the PAM rationale, it allows to verify to what extent current PPM methodologies are capable of obtaining similar declarative model outputs.
Note that AUC cannot be calculated as we only check for the presence/absence of constraints without producing a probability per constraint.
The results in Table \ref{tab:lstms} indicate that the recall and precision are low, which is due to the generation of similar, but slightly different traces, leading to the presence of different constraints compared to the ones actually present in the original trace.
\begin{table}[tb]
	\centering
	\scriptsize
	\resizebox{0.6\textwidth}{!}{
		\begin{tabular}{|r|r|r|r|r|}
			\hline
			\multicolumn{1}{|l|}{\textbf{dataset}} & \multicolumn{1}{l|}{\textbf{window}} & \multicolumn{1}{l|}{\textbf{recall}} & \multicolumn{1}{l|}{\textbf{precision}} & \multicolumn{1}{l|}{\textbf{F-score}} \\
			\hline
			\multicolumn{1}{|l|}{\textbf{BPI12}} & 4     & 0.349 & 0.135 & 0.194 \\
			& 5     & 0.325 & 0.126 & 0.181 \\
			& 10    & 0.288 & 0.363 & 0.321 \\
			\hline
			\multicolumn{1}{|l|}{\textbf{BPI17}} & 4     & 0.291 & 0.179 & 0.221 \\
			& 5     & 0.735 & 0.725 & 0.73 \\
			& 10    & 0.51  & 0.196 & 0.283 \\
			\hline
		\end{tabular}%
	}
	\caption{An overview of the performance of finding constraints in a window of variable sizes at the end of the string as predicted by an LSTM.}
	\label{tab:lstms}
\end{table}

\subsubsection{Predictive accuracy and precision}
\label{sec:accprec}
The results of PAM for both network topologies are included for a fixed number of windows in Table \ref{tab:fixednowin}, and for a fixed window size in Table \ref{tab:fixedwinsize12}.
\begin{table*}
	\centering
	\resizebox{\textwidth}{!}{
		
\begin{tabular}{|c|l|r|r|r|r|r|r|r|r|r|r|}
	\cline{3-12}\multicolumn{1}{r}{} &       & \multicolumn{5}{c|}{\textbf{Convolutional LSTMs}} & \multicolumn{5}{c|}{\textbf{Encoder-decoder LSTMs}} \bigstrut\\
	\hline
	\multicolumn{1}{|l|}{\textbf{event log (\#windows)}} & \textbf{metric} & \multicolumn{1}{c|}{\textbf{mean}} & \multicolumn{1}{c|}{\textbf{std}} & \multicolumn{1}{c|}{\textbf{min}} & \multicolumn{1}{c|}{\textbf{med}} & \multicolumn{1}{c|}{\textbf{max}} & \multicolumn{1}{c|}{\textbf{mean}} & \multicolumn{1}{c|}{\textbf{std}} & \multicolumn{1}{c|}{\textbf{min}} & \multicolumn{1}{c|}{\textbf{med}} & \multicolumn{1}{c|}{\textbf{max}} \bigstrut\\
	\hline
	\multirow{4}[8]{*}{\textbf{BPI 12 (2)}} & \textbf{AP} & \cellcolor[rgb]{ .906,  .902,  .902}\textbf{0.786} & 0.036 & 0.654 & 0.8   & \cellcolor[rgb]{ .906,  .902,  .902}\textbf{0.819} & \multicolumn{1}{c|}{0.737} & \multicolumn{1}{c|}{0.184} & \multicolumn{1}{c|}{0.26} & \multicolumn{1}{c|}{0.806} & \multicolumn{1}{c|}{0.816} \bigstrut\\
	\cline{2-12}      & \textbf{AUC} & \textbf{0.996} & 0.002 & 0.985 & 0.997 & \textbf{0.997} & \multicolumn{1}{c|}{0.954} & \multicolumn{1}{c|}{0.117} & \multicolumn{1}{c|}{0.651} & \multicolumn{1}{c|}{0.997} & \multicolumn{1}{c|}{\textbf{0.997}} \bigstrut\\
	\cline{2-12}      & \textbf{F--score} & \textbf{0.69} & 0.024 & 0.604 & 0.698 & \textbf{0.715} & \multicolumn{1}{c|}{0.672} & \multicolumn{1}{c|}{0.086} & \multicolumn{1}{c|}{0.445} & \multicolumn{1}{c|}{0.704} & \multicolumn{1}{c|}{0.711} \bigstrut\\
	\cline{2-12}      & \textbf{Time} & 16.047 & 6.677 & 9.802 & 13.867 & 37.199 & \multicolumn{1}{c|}{19.732} & \multicolumn{1}{c|}{7.127} & \multicolumn{1}{c|}{10.541} & \multicolumn{1}{c|}{18.617} & \multicolumn{1}{c|}{36.088} \bigstrut\\
	\hline
	\multirow{4}[8]{*}{\textbf{BPI 12 (5)}} & \textbf{AP} & \cellcolor[rgb]{ .906,  .902,  .902}\textbf{0.824} & 0.043 & 0.678 & 0.84  & 0.869 & 0.751 & 0.2   & 0.346 & 0.846 & \cellcolor[rgb]{ .906,  .902,  .902}\textbf{0.871} \bigstrut\\
	\cline{2-12}      & \textbf{AUC} & \textbf{0.995} & 0.014 & 0.906 & 0.998 & \textbf{0.999} & 0.942 & 0.118 & 0.699 & 0.998 & \textbf{0.999} \bigstrut\\
	\cline{2-12}      & \textbf{F--score} & \textbf{0.735} & 0.034 & 0.637 & 0.747 & 0.773 & 0.712 & 0.084 & 0.544 & 0.754 & \textbf{0.776} \bigstrut\\
	\cline{2-12}      & \textbf{Time} & 29.447 & 17.28 & 8.986 & 25.602 & 86.716 & 23.716 & 7.428 & 12.423 & 22.919 & 39.13 \bigstrut\\
	\hline
	\multirow{4}[8]{*}{\textbf{BPI 12 (10)}} & \textbf{AP} & \cellcolor[rgb]{ .906,  .902,  .902}\textbf{0.725} & 0.068 & 0.542 & 0.751 & 0.797 & 0.569 & 0.215 & 0.279 & 0.629 & \cellcolor[rgb]{ .906,  .902,  .902}\textbf{0.803} \bigstrut\\
	\cline{2-12}      & \textbf{AUC} & \textbf{0.988} & 0.022 & 0.873 & 0.995 & \textbf{0.997} & 0.881 & 0.153 & 0.676 & 0.992 & \textbf{0.997} \bigstrut\\
	\cline{2-12}      & \textbf{F--score} & \textbf{0.651} & 0.047 & 0.547 & 0.668 & 0.708 & 0.589 & 0.087 & 0.493 & 0.561 & \textbf{0.71} \bigstrut\\
	\cline{2-12}      & \textbf{Time} & 43.199 & 28.074 & 10.554 & 36.282 & 134.514 & 30.174 & 10.269 & 14.988 & 28.876 & 51.464 \bigstrut\\
	\hline
	\multirow{4}[8]{*}{\textbf{BPI 17 (2)}} & \textbf{AP} & 0.835 & 0.052 & 0.646 & 0.856 & 0.874 & \cellcolor[rgb]{ .906,  .902,  .902}\textbf{0.868} & 0.006 & 0.86  & 0.871 & \cellcolor[rgb]{ .906,  .902,  .902}\textbf{0.875} \bigstrut\\
	\cline{2-12}      & \textbf{AUC} & 0.995 & 0.003 & 0.982 & 0.996 & \textbf{0.997} & \textbf{0.997} & 0     & 0.997 & 0.997 & \textbf{0.997} \bigstrut\\
	\cline{2-12}      & \textbf{F--score} & 0.766 & 0.044 & 0.613 & 0.784 & 0.798 & \textbf{0.791} & 0.006 & 0.783 & 0.793 & \textbf{0.799} \bigstrut\\
	\cline{2-12}      & \textbf{Time} & 43.24 & 19.533 & 23.257 & 40.151 & 103.91 & 48.366 & 17.33 & 26.364 & 45.235 & 93.597 \bigstrut\\
	\hline
	\multirow{4}[8]{*}{\textbf{BPI 17 (5)}} & \textbf{AP} & \cellcolor[rgb]{ .906,  .902,  .902}\textbf{0.887} & 0.03  & 0.775 & 0.901 & \cellcolor[rgb]{ .906,  .902,  .902}\textbf{0.916} & 0.804 & 0.233 & 0.19  & 0.89  & 0.911 \bigstrut\\
	\cline{2-12}      & \textbf{AUC} & \textbf{0.998} & 0.001 & 0.992 & 0.999 & \textbf{0.999} & 0.958 & 0.111 & 0.66  & 0.999 & \textbf{0.999} \bigstrut\\
	\cline{2-12}      & \textbf{F--score} & \textbf{0.814} & 0.028 & 0.714 & 0.825 & \textbf{0.842} & 0.763 & 0.13  & 0.428 & 0.811 & 0.837 \bigstrut\\
	\cline{2-12}      & \textbf{Time} & 115.597 & 70.375 & 30.138 & 93.983 & 334.22 & 302.512 & 43.86 & 232.062 & 312.164 & 388.493 \bigstrut\\
	\hline
	\multirow{4}[8]{*}{\textbf{BPI 17 (10)}} & \textbf{AP} & \cellcolor[rgb]{ .906,  .902,  .902}\textbf{0.84} & 0.045 & 0.673 & 0.853 & \cellcolor[rgb]{ .906,  .902,  .902}\textbf{0.883} & 0.552 & 0.295 & 0.176 & 0.599 & 0.88 \bigstrut\\
	\cline{2-12}      & \textbf{AUC} & \textbf{0.996} & 0.009 & 0.939 & 0.998 & \textbf{0.999} & 0.845 & 0.162 & 0.66  & 0.909 & \textbf{0.999} \bigstrut\\
	\cline{2-12}      & \textbf{F--score} & \textbf{0.758} & 0.037 & 0.634 & 0.768 & \textbf{0.795} & 0.606 & 0.144 & 0.403 & 0.608 & 0.79 \bigstrut\\
	\cline{2-12}      & \textbf{Time} & 422.039 & 73.139 & 188.542 & 411.632 & 707.495 & 390.521 & 176.489 & 156.858 & 414.479 & 581.096 \bigstrut\\
	\hline
\end{tabular}%

			}
	\caption{An overview of the performance of both neural network approaches for a fixed number of windows per trace. The highest average and maximum average precision (AP) is indicated in grey, the other metrics in bold.}
	\label{tab:fixednowin}
\end{table*}
\begin{table*}
	\centering
	\scriptsize
	\resizebox{\textwidth}{!}{

\begin{tabular}{|c|c|l|r|r|r|r|r|r|c|r|r|r|r|r|r|}
\cline{4-9}\cline{11-16}\multicolumn{1}{r}{} & \multicolumn{1}{r}{} &       & \multicolumn{3}{c|}{\textbf{Convolutional LSTMs}} & \multicolumn{3}{c|}{\textbf{Encoder-decoder LSTMs}} &       & \multicolumn{3}{c|}{\textbf{Convolutional LSTMs}} & \multicolumn{3}{c|}{\textbf{Encoder-decoder LSTMs}} \bigstrut\\
\hline
\multicolumn{1}{|l|}{\textbf{event log}} & \multicolumn{1}{l|}{\textbf{\#win.}} & \textbf{metric} & \multicolumn{1}{c|}{\textbf{mean}} & \multicolumn{1}{c|}{\textbf{min}} & \multicolumn{1}{c|}{\textbf{max}} & \multicolumn{1}{c|}{\textbf{mean}} & \multicolumn{1}{c|}{\textbf{min}} & \multicolumn{1}{c|}{\textbf{max}} & \multicolumn{1}{l|}{\textbf{Event log}} & \multicolumn{1}{c|}{\textbf{mean}} & \multicolumn{1}{c|}{\textbf{min}} & \multicolumn{1}{c|}{\textbf{max}} & \multicolumn{1}{c|}{\textbf{mean}} & \multicolumn{1}{c|}{\textbf{min}} & \multicolumn{1}{c|}{\textbf{max}} \bigstrut\\
\hline
\multirow{15}[30]{*}{\textbf{BPI 12 (2)}} & \multirow{3}[6]{*}{\textbf{6--10}} & \textbf{AP} & \cellcolor[rgb]{ .906,  .902,  .902}\textbf{0.973} & 0.799 & \cellcolor[rgb]{ .906,  .902,  .902}\textbf{0.992} & 0.574 & 0.005 & 0.982 & \multirow{15}[30]{*}{\textbf{BPI 17 (2)}} & \cellcolor[rgb]{ .906,  .902,  .902}\textbf{0.96} & 0.559 & \cellcolor[rgb]{ .906,  .902,  .902}\textbf{1} & 0.819 & 0.109 & 0.999 \bigstrut\\
\cline{3-9}\cline{11-16}      &       & \textbf{AUC} & \textbf{0.997} & 0.91  & \textbf{1} & 0.669 & 0.49  & \textbf{1} &       & \textbf{0.964} & 0.463 & \textbf{1} & 0.842 & 0.499 & \textbf{1} \bigstrut\\
\cline{3-9}\cline{11-16}      &       & \textbf{Time} & 8.82  & 4.237 & 16.33 & 4.015 & 1.923 & 6.173 &       & 16.732 & 7.137 & 36.557 & 7.216 & 3.629 & 11.189 \bigstrut\\
\cline{2-9}\cline{11-16}      & \multirow{3}[6]{*}{\textbf{11--15}} & \textbf{AP} & \cellcolor[rgb]{ .906,  .902,  .902}\textbf{0.966} & 0.832 & \cellcolor[rgb]{ .906,  .902,  .902}\textbf{0.996} & 0.71  & 0.234 & 0.977 &       & \cellcolor[rgb]{ .906,  .902,  .902}\textbf{0.963} & 0.778 & \cellcolor[rgb]{ .906,  .902,  .902}\textbf{1} & 0.724 & 0.301 & 0.999 \bigstrut\\
\cline{3-9}\cline{11-16}      &       & \textbf{AUC} & \textbf{0.999} & 0.978 & \textbf{1} & 0.673 & 0.491 & 0.998 &       & \textbf{0.981} & 0.886 & \textbf{1} & 0.641 & 0.499 & 0.998 \bigstrut\\
\cline{3-9}\cline{11-16}      &       & \textbf{Time} & 29.167 & 8.638 & 230.284 & 8.876 & 4.715 & 13.645 &       & 141.21 & 51.816 & 376.791 & 56.018 & 40.244 & 81.215 \bigstrut\\
\cline{2-9}\cline{11-16}      & \multirow{3}[6]{*}{\textbf{16--20}} & \textbf{AP} & \cellcolor[rgb]{ .906,  .902,  .902}\textbf{0.934} & 0.629 & \cellcolor[rgb]{ .906,  .902,  .902}\textbf{0.973} & 0.649 & 0.303 & 0.961 &       & \cellcolor[rgb]{ .906,  .902,  .902}\textbf{0.979} & 0.813 & \cellcolor[rgb]{ .906,  .902,  .902}\textbf{1} & 0.715 & 0.168 & 0.998 \bigstrut\\
\cline{3-9}\cline{11-16}      &       & \textbf{AUC} & \textbf{0.994} & 0.923 & \textbf{1} & 0.531 & 0.496 & 0.999 &       & \textbf{0.993} & 0.906 & \textbf{1} & 0.654 & 0.495 & 0.998 \bigstrut\\
\cline{3-9}\cline{11-16}      &       & \textbf{Time} & 25.853 & 9.197 & 68.545 & 9.313 & 4.861 & 14.31 &       & 130.98 & 45.208 & 349.506 & 47.382 & 25.15 & 71.148 \bigstrut\\
\cline{2-9}\cline{11-16}      & \multirow{3}[6]{*}{\textbf{21--25}} & \textbf{AP} & \cellcolor[rgb]{ .906,  .902,  .902}\textbf{0.891} & 0.025 & \cellcolor[rgb]{ .906,  .902,  .902}\textbf{0.948} & 0.657 & 0.459 & 0.724 &       & \cellcolor[rgb]{ .906,  .902,  .902}\textbf{0.949} & 0.032 & \cellcolor[rgb]{ .906,  .902,  .902}\textbf{1} & 0.645 & 0.255 & 0.869 \bigstrut\\
\cline{3-9}\cline{11-16}      &       & \textbf{AUC} & \textbf{0.993} & 0.804 & \textbf{1} & 0.5   & 0.5   & 0.5   &       & \textbf{0.981} & 0.546 & \textbf{1} & 0.5   & 0.496 & 0.5 \bigstrut\\
\cline{3-9}\cline{11-16}      &       & \textbf{Time} & 18.425 & 6.536 & 49.744 & 6.507 & 3.325 & 9.953 &       & 120.95 & 40.925 & 325.313 & 42.942 & 22.604 & 65.154 \bigstrut\\
\cline{2-9}\cline{11-16}      & \multirow{3}[6]{*}{\textbf{26--30}} & \textbf{AP} & \cellcolor[rgb]{ .906,  .902,  .902}\textbf{0.906} & 0.838 & \cellcolor[rgb]{ .906,  .902,  .902}\textbf{0.931} & 0.659 & 0.327 & 0.75  &       & \cellcolor[rgb]{ .906,  .902,  .902}\textbf{0.945} & 0.028 & \cellcolor[rgb]{ .906,  .902,  .902}\textbf{0.994} & 0.673 & 0.325 & 0.759 \bigstrut\\
\cline{3-9}\cline{11-16}      &       & \textbf{AUC} & \textbf{0.996} & 0.989 & \textbf{0.999} & 0.5   & 0.5   & 0.5   &       & \textbf{0.97} & 0.402 & \textbf{1} & 0.5   & 0.5   & 0.5 \bigstrut\\
\cline{3-9}\cline{11-16}      &       & \textbf{Time} & 43.557 & 4.037 & 1160.43 & 3.935 & 2.002 & 6.1   &       & 87.738 & 28.326 & 233.279 & 29.611 & 15.747 & 45.045 \bigstrut\\
\hline
\multirow{15}[30]{*}{\textbf{BPI 12 (5)}} & \multirow{3}[6]{*}{\textbf{3--4}} & \textbf{AP} & \cellcolor[rgb]{ .906,  .902,  .902}\textbf{0.941} & 0.853 & \cellcolor[rgb]{ .906,  .902,  .902}\textbf{0.953} & 0.877 & 0.564 & 0.948 & \multirow{15}[30]{*}{\textbf{BPI 17 (5)}} & \cellcolor[rgb]{ .906,  .902,  .902}\textbf{0.946} & 0.52  & \cellcolor[rgb]{ .906,  .902,  .902}\textbf{0.992} & 0.945 & 0.343 & 0.991 \bigstrut\\
\cline{3-9}\cline{11-16}      &       & \textbf{AUC} & \textbf{0.999} & 0.992 & \textbf{1} & 0.973 & 0.817 & \textbf{1} &       & 0.977 & 0.74  & \textbf{1} & \textbf{0.981} & 0.711 & \textbf{1} \bigstrut\\
\cline{3-9}\cline{11-16}      &       & \textbf{Time} & 4.547 & 2.229 & 7.939 & 2.487 & 1.376 & 3.863 &       & 7.944 & 3.803 & 16.159 & 4.318 & 2.232 & 6.791 \bigstrut\\
\cline{2-9}\cline{11-16}      & \multirow{3}[6]{*}{\textbf{5--6}} & \textbf{AP} & \cellcolor[rgb]{ .906,  .902,  .902}\textbf{0.922} & 0.849 & \cellcolor[rgb]{ .906,  .902,  .902}\textbf{0.948} & 0.799 & 0.46  & 0.933 &       & \cellcolor[rgb]{ .906,  .902,  .902}\textbf{0.972} & 0.616 & \cellcolor[rgb]{ .906,  .902,  .902}\textbf{0.991} & 0.913 & 0.402 & \cellcolor[rgb]{ .906,  .902,  .902}\textbf{0.991} \bigstrut\\
\cline{3-9}\cline{11-16}      &       & \textbf{AUC} & \textbf{0.999} & 0.997 & \textbf{1} & 0.962 & 0.773 & 0.999 &       & \textbf{0.992} & 0.802 & \textbf{1} & 0.975 & 0.803 & \textbf{1} \bigstrut\\
\cline{3-9}\cline{11-16}      &       & \textbf{Time} & 11.44 & 4.454 & 28.208 & 5.078 & 2.475 & 8.016 &       & 65.707 & 23.683 & 165.441 & 28.821 & 14.266 & 44.884 \bigstrut\\
\cline{2-9}\cline{11-16}      & \multirow{3}[6]{*}{\textbf{7--8}} & \textbf{AP} & \cellcolor[rgb]{ .906,  .902,  .902}\textbf{0.873} & 0.665 & \cellcolor[rgb]{ .906,  .902,  .902}\textbf{0.936} & 0.795 & 0.507 & 0.899 &       & \cellcolor[rgb]{ .906,  .902,  .902}\textbf{0.962} & 0.818 & \cellcolor[rgb]{ .906,  .902,  .902}\textbf{0.981} & 0.813 & 0.362 & 0.973 \bigstrut\\
\cline{3-9}\cline{11-16}      &       & \textbf{AUC} & \textbf{0.996} & 0.976 & \textbf{0.999} & 0.974 & 0.781 & \textbf{0.999} &       & \textbf{0.998} & 0.97  & \textbf{1} & 0.947 & 0.775 & 1 \bigstrut\\
\cline{3-9}\cline{11-16}      &       & \textbf{Time} & 12.307 & 4.385 & 31.236 & 5.137 & 2.508 & 7.873 &       & 62.294 & 22.007 & 157.082 & 24.881 & 12.399 & 37.663 \bigstrut\\
\cline{2-9}\cline{11-16}      & \multirow{3}[6]{*}{\textbf{9--10}} & \textbf{AP} & \cellcolor[rgb]{ .906,  .902,  .902}\textbf{0.8} & 0.533 & \cellcolor[rgb]{ .906,  .902,  .902}\textbf{0.909} & 0.661 & 0.293 & 0.846 &       & \cellcolor[rgb]{ .906,  .902,  .902}\textbf{0.927} & 0.692 & \cellcolor[rgb]{ .906,  .902,  .902}\textbf{0.98} & 0.7   & 0.251 & 0.964 \bigstrut\\
\cline{3-9}\cline{11-16}      &       & \textbf{AUC} & \textbf{0.994} & 0.954 & \textbf{0.999} & 0.918 & 0.72  & 0.998 &       & \textbf{0.992} & 0.851 & \textbf{1} & 0.905 & 0.763 & \textbf{1} \bigstrut\\
\cline{3-9}\cline{11-16}      &       & \textbf{Time} & 8.76  & 3.064 & 22.387 & 3.518 & 1.778 & 5.575 &       & 57.955 & 19.484 & 146.974 & 22.138 & 12.16 & 33.098 \bigstrut\\
\cline{2-9}\cline{11-16}      & \multirow{3}[6]{*}{\textbf{11--12}} & \textbf{AP} & \cellcolor[rgb]{ .906,  .902,  .902}\textbf{0.73} & 0.511 & \cellcolor[rgb]{ .906,  .902,  .902}\textbf{0.798} & 0.61  & 0.239 & 0.768 &       & \cellcolor[rgb]{ .906,  .902,  .902}\textbf{0.891} & 0.071 & \cellcolor[rgb]{ .906,  .902,  .902}\textbf{0.975} & 0.637 & 0.167 & 0.966 \bigstrut\\
\cline{3-9}\cline{11-16}      &       & \textbf{AUC} & \textbf{0.988} & 0.848 & \textbf{0.997} & 0.907 & 0.715 & 0.996 &       & \textbf{0.984} & 0.661 & \textbf{1} & 0.881 & 0.763 & \textbf{1} \bigstrut\\
\cline{3-9}\cline{11-16}      &       & \textbf{Time} & 5.608 & 1.927 & 14.176 & 2.152 & 1.117 & 3.371 &       & 39.968 & 13.066 & 102.656 & 14.626 & 7.361 & 22.494 \bigstrut\\
\hline
\multirow{15}[30]{*}{\textbf{BPI 12 (10)}} & \multirow{3}[6]{*}{\textbf{2}} & \textbf{AP} & \cellcolor[rgb]{ .906,  .902,  .902}\textbf{0.922} & 0.887 & \cellcolor[rgb]{ .906,  .902,  .902}\textbf{0.957} & 0.858 & 0.377 & 0.947 & \multirow{15}[30]{*}{\textbf{BPI 17 (10)}} & \cellcolor[rgb]{ .906,  .902,  .902}\textbf{0.979} & 0.968 & \cellcolor[rgb]{ .906,  .902,  .902}\textbf{0.989} & 0.956 & 0.769 & 0.988 \bigstrut\\
\cline{3-9}\cline{11-16}      &       & \textbf{AUC} & \textbf{0.998} & 0.989 & \textbf{0.999} & 0.965 & 0.728 & \textbf{0.999} &       & \textbf{0.999} & 0.998 & \textbf{1} & 0.988 & 0.908 & \textbf{1} \bigstrut\\
\cline{3-9}\cline{11-16}      &       & \textbf{Time} & 2.423 & 1.449 & 3.78  & 1.748 & 1.122 & 2.916 &       & 4.299 & 2.469 & 7.444 & 2.957 & 1.754 & 4.751 \bigstrut\\
\cline{2-9}\cline{11-16}      & \multirow{3}[6]{*}{\textbf{3}} & \textbf{AP} & \cellcolor[rgb]{ .906,  .902,  .902}\textbf{0.827} & 0.779 & 0.854 & 0.787 & 0.342 & \cellcolor[rgb]{ .906,  .902,  .902}\textbf{0.858} &       & 0.921 & 0.884 & 0.938 & \cellcolor[rgb]{ .906,  .902,  .902}\textbf{0.93} & 0.913 & \cellcolor[rgb]{ .906,  .902,  .902}\textbf{0.939} \bigstrut\\
\cline{3-9}\cline{11-16}      &       & \textbf{AUC} & \textbf{0.997} & 0.994 & 0.998 & 0.979 & 0.705 & \textbf{0.999} &       & 0.998 & 0.977 & \textbf{0.999} & \textbf{0.999} & 0.995 & \textbf{0.999} \bigstrut\\
\cline{3-9}\cline{11-16}      &       & \textbf{Time} & 5.585 & 2.679 & 12.431 & 3.404 & 1.815 & 5.906 &       & 32.042 & 14.555 & 73.366 & 19.018 & 9.618 & 34.132 \bigstrut\\
\cline{2-9}\cline{11-16}      & \multirow{3}[6]{*}{\textbf{4}} & \textbf{AP} & \cellcolor[rgb]{ .906,  .902,  .902}\textbf{0.821} & 0.711 & \cellcolor[rgb]{ .906,  .902,  .902}\textbf{0.868} & 0.727 & 0.261 & 0.851 &       & \cellcolor[rgb]{ .906,  .902,  .902}\textbf{0.9} & 0.854 & \cellcolor[rgb]{ .906,  .902,  .902}\textbf{0.92} & 0.859 & 0.4   & 0.914 \bigstrut\\
\cline{3-9}\cline{11-16}      &       & \textbf{AUC} & \textbf{0.997} & 0.99  & \textbf{0.999} & 0.96  & 0.698 & 0.998 &       & \textbf{0.999} & 0.997 & \textbf{0.999} & 0.982 & 0.738 & \textbf{0.999} \bigstrut\\
\cline{3-9}\cline{11-16}      &       & \textbf{Time} & 6.08  & 2.567 & 14.182 & 3.253 & 1.761 & 5.269 &       & 29.948 & 12.292 & 71.597 & 15.735 & 8.166 & 27.556 \bigstrut\\
\cline{2-9}\cline{11-16}      & \multirow{3}[6]{*}{\textbf{5}} & \textbf{AP} & \cellcolor[rgb]{ .906,  .902,  .902}\textbf{0.758} & 0.652 & \cellcolor[rgb]{ .906,  .902,  .902}\textbf{0.819} & 0.681 & 0.166 & 0.783 &       & \cellcolor[rgb]{ .906,  .902,  .902}\textbf{0.9} & 0.754 & \textbf{0.924} & 0.833 & 0.326 & 0.913 \bigstrut\\
\cline{3-9}\cline{11-16}      &       & \textbf{AUC} & \textbf{0.994} & 0.987 & \textbf{0.997} & 0.974 & 0.665 & \textbf{0.997} &       & \textbf{0.997} & 0.954 & \textbf{0.999} & 0.982 & 0.734 & \textbf{0.999} \bigstrut\\
\cline{3-9}\cline{11-16}      &       & \textbf{Time} & 4.352 & 1.733 & 10.35 & 2.206 & 1.064 & 3.546 &       & 27.997 & 10.725 & 67.197 & 13.343 & 6.21  & 22.833 \bigstrut\\
\cline{2-9}\cline{11-16}      & \multirow{3}[6]{*}{\textbf{6}} & \textbf{AP} & \cellcolor[rgb]{ .906,  .902,  .902}\textbf{0.688} & 0.597 & 0.772 & 0.625 & 0.189 & \cellcolor[rgb]{ .906,  .902,  .902}\textbf{0.779} &       & \cellcolor[rgb]{ .906,  .902,  .902}\textbf{0.897} & 0.836 & \cellcolor[rgb]{ .906,  .902,  .902}\textbf{0.922} & 0.8   & 0.147 & 0.908 \bigstrut\\
\cline{3-9}\cline{11-16}      &       & \textbf{AUC} & \textbf{0.991} & 0.978 & 0.995 & 0.951 & 0.653 & \textbf{0.997} &       & \textbf{0.998} & 0.996 & \textbf{0.999} & 0.981 & 0.73  & \textbf{0.999} \bigstrut\\
\cline{3-9}\cline{11-16}      &       & \textbf{Time} & 2.898 & 1.131 & 7.01  & 1.356 & 0.692 & 2.144 &       & 20.053 & 7.425 & 50.366 & 8.973 & 4.414 & 14.846 \bigstrut\\
\hline
\end{tabular}%

	}
	\caption{An overview of the performance of both neural network approaches for a fixed window size (in brackets after event log name) trained over different subsets of the 2012 and 2017 BPI Challenge data containing different numbers of windows. The highest average and maximum AP is indicated in grey, the other metrics in bold.}
	\label{tab:fixedwinsize12}
\end{table*}

For a fixed number of windows, we see that convolutional LSTMs typically perform better on average, with the maximum performance (i.e. the hyperparameter combination producing the best result) is relatively similar for both network topologies.
Overall, the results for the BPI 17 log, with its longer traces are higher, with consistently high AUC, and AP up to 92\% for 5 windows per trace. 
For the BPI 12 log the results are lower, with more windows (10) having the lowest AP, probably due to the small number of activities present in the windows.
In these scenarios (number of windows at 10), the convolutional LSTMs have much more reliable results.
Note also that there is significantly more training data available for the BPI 17 log.
It is interesting to note that even when dividing a trace in 2 (fixed number of windows of 2) results in an AP of up to 82\% and 88\% for the 12/17 logs respectively, meaning that even from 1 window the LSTMs can learn what constraints will be present in the second half of the trace.
For a fixed window size, again we see better and more consistent results for the convolutional LSTMs, which maximal performance again being close with encoder-decoder LSTMs.
For window sizes 5-10, high average precision up to 100\% can be achieved, with results being lower for longer traces.
This is likely due to the fact that fewer examples are available to train the networks, as this is especially prominent for the BPI 12 log where few longer traces are available.
Again, the performance for the BPI 17 logs is stronger, with average precision reaching 90\% on average consistently.

These results show a good balance between AUC and AP, meaning the networks are capable of predicting correct true positives in a sparse environment without generating too many false positives as evidenced by the high levels of AP.
Overall, the convolutional LSTMs perform better, but often come at a run-time cost.
In general, the scores are at least comparable or better to the AUC reported for the 2012 and 2017 log as included in \cite{DBLP:journals/corr/TeinemaaDRM17} for the prediction of single objectives.

In comparison with the baseline from Table \ref{tab:lstms}, the network topologies used for PAM are better capable of creating appropriate embeddings based on the constraints that are more informative towards prediction.
As illustrated in Section \ref{sec:conint}, both positional constraints such as \textit{absence}, but also \textit{precedence} contribute towards the overall result, meaning that sequential relationships (potentially over long distance dependencies captured by LSTMs) are adequately captured in the convolutional layer. These results clearly show the added value of the PAM architecture with respect to the state-of-the-art in the field. 

\subsubsection{Neural network parameterisation} 
An overview of the impact of the network parameterisation on average precision is not included but can be found for both network approaches at \url{https://github.com/JohannesDeSmedt/processes-as-movies}.
Overall, the number of ConvLSTM layers used, the size of the filters, and kernel size have little impact on the average precision for a fixed number of windows.
Hence, the intuition of Section \ref{sec:4network} cannot necessarily be applied, potentially due to the high overlap and slow changes in tensors between time steps even when few time steps are used for longer traces.
The kernel size does result in slightly different results for BPI 17 with 10 windows per trace.
For fixed window sizes, the results also remain relatively stable, with a single drop in AP typically for higher filter sizes in combination with higher kernel sizes, however, the occasional spikes do not follow a specific trend. 
For more windows per trace, regardless of the window size, the results become more varied, possibly due to the fewer examples available.
The results for window size 10 contain the least variance.

For encoder-decoder LSTMs, the impact of the feature dimensionality used does seem to have a strong effect.
The higher the dimensionality, the lower the average precision for 5 and 10 windows per trace for a fixed number of windows.
This could potentially indicate that the networks overfit quicker.
For the fixed window length, the results are relatively unstable with again lower AP for higher dimensions.
This is especially apparent for window size 2.
Given that this approach is more similar to other LSTM-based approaches, it might indicate that the ConvLSTMs might even excel for single next-in-sequence activity prediction.

Overall, we can notice that the results of the convolutional LSTMs are much more stable over the full parameter space, which is in line with the standard deviations from Tables \ref{tab:fixednowin}--\ref{tab:fixedwinsize12}.

\subsection{Constraints and interpretation}
\label{sec:conint}
Given that PAM resorts to predicting various types of constraints at once in the output tensor, it is possible to retrieve the evaluation metrics per constraint type as well.
In Table \ref{tab:con-fixednowin}, the results for the fixed number of windows approach are shown, and in Tables \ref{tab:con-fixedwinlen-ed} and \ref{tab:con-fixedwinlen-conv} the results for the fixed window length approach are shown for the maximum results from Section \ref{sec:accprec} for the BPI 17 log (the BPI 12 log produced similar results, which can be found for encoder-decoder LSTMs \href{https://github.com/JohannesDeSmedt/processes-as-movies/blob/master/results/encoder_decoder_constraints_fixed_size_bpi12}{here} and ConvLSTMs \href{https://github.com/JohannesDeSmedt/processes-as-movies/blob/master/results/convlstm_constraints_fixed_size_bpi12}{here}).
Both the number of constraints predicted to be present, and their average precision is included as AUC is generally high at the same levels as in Tables \ref{tab:con-fixednowin}-\ref{tab:fixedwinsize12}.

Firstly, it is apparent that the average precision is strongly dependent on the number of constraints present.
Hence, infrequently occurring constraints such as \textit{alternate precedence/response} (which are only considered by iBCM in case more than 1 occurrence of the consequent activities is present) have very low precision.
\textit{Absence}, on the other hand, often dominates the overall proportion of constraints present.
Its presence is high as it checks for all non-existing occurrences, which can be plenty in the case of shorter windows where fewer activities of the full activity set appear.
The average precision of \textit{absence} is typically very high at levels close to 99\% for both a fixed number of windows and fixed window size (except for longer traces with fixed window length 2, although this might again be due to fewer observations available).
This seems to drive the average precision of the whole tensor prediction up.

For a fixed number of windows, \textit{precedence}, \textit{response}, \textit{not succession}, and \textit{co-existence} have a noticeably higher proportion as well.
The average precision is typically reasonably high (70-85\%) for the BPI 17 log, while lower for BPI 12 (45-65\%).
There seems to be more of such constraints present in the BPI 17 log as well, with higher proportions for these constraints, meaning there are more examples to learn from.
While not present/predicted as often, the unary constraints \textit{exactly (2)/existence/init/last} all achieve very high average precision scores (80-90\%), although this is less the case for \textit{exactly (2)} for the BPI 17 log, and for a higher number of windows for BPI 12.
In general, this constraint occurs less often than the other unary constraints, so again, the number of training samples is scarcer. 
There are no noticeable differences between both neural network topologies.

For a fixed window size, the dominance of \textit{absence} is even stronger, with other constraints making up very low proportions of the other constraints.
This is mostly due to the shorter windows, in which fewer binary constraints can manifest but more of the rest of the activity set is not present. 
Some constraints can also not be present (e.g. \textit{existence (3)}).
For shorter window lengths (2), all constraints have high average precision, which is likely caused by the higher number of steps that are fed to the LSTM compared to other settings.
For the very long traces (21--25, 26--30 windows), the average precision slacks off for encoder-decoder LSTMs and only absence is predicted with reasonable precision, possibly due to fewer traces available for training or too few constraints present in general.
For longer window lengths (5/10), constraints with a higher proportion perform better, especially for the BPI 17 log, although again this trails off for longer traces.
For a window size of 5, binary constraints perform around 50\% for BPI 12, and 80-90\% for BPI 17.
For a window size of 10, the binary constraints achieve 80\% AP and up for BPI 12, and 95\% for BPI 17.
Exactly and Init/Last perform at +85\% for window sizes 5 and 10 for BPI 12, and 99\% scores for BPI 17.

Again, there seems to be little difference in performance per constraint between the best-performing models for either convolutional and encoder-decoder LSTMs, however, for a window size of 2 in longer traces there are more positive predictions by the convolutional LSTM over all constraints, with high precision rates.
The same is noticeable for a window length of 5 and longer traces to some extent.
Convolutional LSTMs seem to be better capable of making these fine-granular predictions.

\begin{table*}
	\centering
	\scriptsize
	\resizebox{\textwidth}{!}{
		
\begin{tabular}{|l|r|r|r|r|r|r|r|r|r|r|r|r|}
	\cline{2-13}\multicolumn{1}{r|}{} & \multicolumn{6}{c|}{\textbf{Convolutional LSTMs}} & \multicolumn{6}{c|}{\textbf{Encoder-decoder LSTMs}} \bigstrut\\
	\hline
	\textbf{constraint} & \multicolumn{1}{l|}{\textbf{count}} & \multicolumn{1}{l|}{\textbf{ap}} & \multicolumn{1}{l|}{\textbf{count}} & \multicolumn{1}{l|}{\textbf{ap}} & \multicolumn{1}{l|}{\textbf{count}} & \multicolumn{1}{l|}{\textbf{ap}} & \multicolumn{1}{l|}{\textbf{count}} & \multicolumn{1}{l|}{\textbf{ap}} & \multicolumn{1}{l|}{\textbf{count}} & \multicolumn{1}{l|}{\textbf{ap}} & \multicolumn{1}{l|}{\textbf{count}} & \multicolumn{1}{l|}{\textbf{ap}} \bigstrut\\
	\hline
	\textbf{BPI 12} & \multicolumn{2}{c|}{\textbf{2}} & \multicolumn{2}{c|}{\textbf{5}} & \multicolumn{2}{c|}{\textbf{10}} & \multicolumn{2}{c|}{\textbf{2}} & \multicolumn{2}{c|}{\textbf{5}} & \multicolumn{2}{c|}{\textbf{10}} \bigstrut\\
	\hline
	\textbf{Prec.} & \cellcolor[rgb]{ .82,  .82,  .82}    23,544  & \cellcolor[rgb]{ .776,  .776,  .776}0.612 & \cellcolor[rgb]{ .867,  .867,  .867}    11,199  & \cellcolor[rgb]{ .761,  .761,  .761}0.641 & \cellcolor[rgb]{ .89,  .89,  .89}       9,021  & \cellcolor[rgb]{ .827,  .827,  .827}0.505 & \cellcolor[rgb]{ .816,  .816,  .816}    24,034  & \cellcolor[rgb]{ .773,  .773,  .773}0.620 & \cellcolor[rgb]{ .871,  .871,  .871}    10,962  & \cellcolor[rgb]{ .792,  .792,  .792}0.622 & \cellcolor[rgb]{ .89,  .89,  .89}       9,066  & \cellcolor[rgb]{ .824,  .824,  .824}0.513 \bigstrut\\
	\hline
	\textbf{Alt. prec} & \cellcolor[rgb]{ .949,  .949,  .949}          228  & \cellcolor[rgb]{ .949,  .949,  .949}0.062 & \cellcolor[rgb]{ .949,  .949,  .949}            12  & \cellcolor[rgb]{ .949,  .949,  .949}0.019 &                6  & 0.009 & \cellcolor[rgb]{ .949,  .949,  .949}          259  & \cellcolor[rgb]{ .949,  .949,  .949}0.064 & \cellcolor[rgb]{ .949,  .949,  .949}            21  & 0.051 &             12  & 0.013 \bigstrut\\
\hline
\end{tabular}%

	}
	\caption{An overview of the number of constraints present, and their average precision for the highest scoring result from Table \ref{tab:fixednowin} for a fixed window size. The grey scale indicates overall proportion per column/number of windows/data set combination overall all constraints.}
	\label{tab:con-fixednowin}
\end{table*}
\begin{table}
	\centering
	\resizebox{0.65\textwidth}{!}{

\begin{tabular}{|l|r|r|r|r|r|r|r|r|r|r|}
\cline{2-11}\multicolumn{1}{r|}{} & \multicolumn{10}{c|}{\textbf{Encoder-decoder LSTMs - BPI 17}} \bigstrut\\
\cline{2-11}\multicolumn{1}{r|}{} & \multicolumn{10}{c|}{\textbf{Window length 2}} \bigstrut\\
\cline{2-11}\multicolumn{1}{r|}{} & \multicolumn{2}{c|}{\textbf{6--10}} & \multicolumn{2}{c|}{\textbf{11--15}} & \multicolumn{2}{c|}{\textbf{16--20}} & \multicolumn{2}{c|}{\textbf{21--25}} & \multicolumn{2}{c|}{\textbf{26--30}} \bigstrut[t]\\
\textbf{constraint} & \multicolumn{1}{l|}{\textbf{count}} & \multicolumn{1}{l|}{\textbf{ap}} & \multicolumn{1}{l|}{\textbf{count}} & \multicolumn{1}{l|}{\textbf{ap}} & \multicolumn{1}{l|}{\textbf{count}} & \multicolumn{1}{l|}{\textbf{ap}} & \multicolumn{1}{l|}{\textbf{count}} & \multicolumn{1}{l|}{\textbf{ap}} & \multicolumn{1}{l|}{\textbf{count}} & \multicolumn{1}{l|}{\textbf{ap}} \bigstrut[b]\\
\hline
\textbf{Prec.} & \cellcolor[rgb]{ .937,  .937,  .937}397 & \cellcolor[rgb]{ .655,  .655,  .655}0.994 & \cellcolor[rgb]{ .945,  .945,  .945}1077 & \cellcolor[rgb]{ .655,  .655,  .655}0.992 & \cellcolor[rgb]{ .945,  .945,  .945}684 & \cellcolor[rgb]{ .663,  .663,  .663}0.962 & \cellcolor[rgb]{ .945,  .945,  .945}515 & \cellcolor[rgb]{ .949,  .949,  .949}0.001 & \cellcolor[rgb]{ .945,  .945,  .945}269 & \cellcolor[rgb]{ .949,  .949,  .949}0.001 \bigstrut\\
\hline
\textbf{Alt. prec} & \cellcolor[rgb]{ .949,  .949,  .949}0 & \cellcolor[rgb]{ .949,  .949,  .949}0.000 & \cellcolor[rgb]{ .949,  .949,  .949}0 & \cellcolor[rgb]{ .949,  .949,  .949}0.000 & \cellcolor[rgb]{ .949,  .949,  .949}0 & \cellcolor[rgb]{ .949,  .949,  .949}0.000 & \cellcolor[rgb]{ .949,  .949,  .949}0 & \cellcolor[rgb]{ .949,  .949,  .949}0.000 & \cellcolor[rgb]{ .949,  .949,  .949}0 & \cellcolor[rgb]{ .949,  .949,  .949}0.000 \bigstrut\\
\hline
\end{tabular}
}
	\caption{An overview of the number of constraints present, and their average precision for the highest scoring result from Table \ref{tab:fixedwinsize12} for a fixed window size and encoder-decoder LSTMs.}
	\label{tab:con-fixedwinlen-ed}
\end{table}
\begin{table}
	\centering
	\resizebox{0.65\textwidth}{!}{

\begin{tabular}{|l|r|r|r|r|r|r|r|r|r|r|}
\cline{2-11}\multicolumn{1}{r|}{} & \multicolumn{10}{c|}{\textbf{Convolutional LSTMs - BPI 17}} \bigstrut\\
\cline{2-11}\multicolumn{1}{r|}{} & \multicolumn{10}{c|}{\textbf{Window length 2}} \bigstrut\\
\cline{2-11}\multicolumn{1}{r|}{} & \multicolumn{2}{c|}{\textbf{6--10}} & \multicolumn{2}{c|}{\textbf{11--15}} & \multicolumn{2}{c|}{\textbf{16--20}} & \multicolumn{2}{c|}{\textbf{21--25}} & \multicolumn{2}{c|}{\textbf{26--30}} \bigstrut[t]\\
\textbf{constraint} & \multicolumn{1}{l|}{\textbf{count}} & \multicolumn{1}{l|}{\textbf{ap}} & \multicolumn{1}{l|}{\textbf{count}} & \multicolumn{1}{l|}{\textbf{ap}} & \multicolumn{1}{l|}{\textbf{count}} & \multicolumn{1}{l|}{\textbf{ap}} & \multicolumn{1}{l|}{\textbf{count}} & \multicolumn{1}{l|}{\textbf{ap}} & \multicolumn{1}{l|}{\textbf{count}} & \multicolumn{1}{l|}{\textbf{ap}} \bigstrut[b]\\
\hline
\textbf{Prec.} & \cellcolor[rgb]{ .937,  .937,  .937}402 & \cellcolor[rgb]{ .655,  .655,  .655}0.989 & \cellcolor[rgb]{ .945,  .945,  .945}1096 & \cellcolor[rgb]{ .655,  .655,  .655}0.996 & \cellcolor[rgb]{ .945,  .945,  .945}659 & \cellcolor[rgb]{ .655,  .655,  .655}0.989 & \cellcolor[rgb]{ .945,  .945,  .945}496 & \cellcolor[rgb]{ .655,  .655,  .655}0.989 & \cellcolor[rgb]{ .945,  .945,  .945}281 & \cellcolor[rgb]{ .733,  .733,  .733}0.736 \bigstrut\\
\hline
\textbf{Alt. prec} & \cellcolor[rgb]{ .949,  .949,  .949}0 & \cellcolor[rgb]{ .949,  .949,  .949}0.000 & \cellcolor[rgb]{ .949,  .949,  .949}0 & \cellcolor[rgb]{ .949,  .949,  .949}0.000 & \cellcolor[rgb]{ .949,  .949,  .949}0 & \cellcolor[rgb]{ .949,  .949,  .949}0.000 & \cellcolor[rgb]{ .949,  .949,  .949}0 & \cellcolor[rgb]{ .949,  .949,  .949}0.000 & \cellcolor[rgb]{ .949,  .949,  .949}0 & \cellcolor[rgb]{ .949,  .949,  .949}0.000 \bigstrut\\
\hline
\end{tabular}
	}
	\caption{An overview of the number of constraints present, and their average precision for the highest scoring result from Table \ref{tab:fixedwinsize12} for a fixed window size and CONVLSTMs.}
	\label{tab:con-fixedwinlen-conv}
\end{table}

\subsection{Discussion}
Distilling all results, it is apparent that PAM with convolutional LSTMs is capable of achieving high predictive accuracy and precision over the high-dimensional output consisting of activity pair-constraint information.
Especially for a log with longer traces (BPI 17), the approach achieves results over 90\% precision.
Very granular input of short fixed window sizes (length 2 or 5), are predicted very well, while on the other side of the spectrum, precision of 70-80\% can still be achieved even when dividing a trace in 2 (fixed number of windows at 2) using only the first half of the trace to predict the presence of constraints in the second half.
Overall, the performance of convolutional neural networks is better and more stable, but come at a runtime cost. However, given that the results do not vary much according to the parameters, even fast-running models can achieve good results.
Given that the performance for short windows is strong, PAM can be used for next-in-sequence prediction as well.
\textit{Existence} or \textit{absence} constraints can tell what activities will be executed next.
Essentially, as discussed before, using only one of these constraints would create a $|A|\times|A|\times1$ tensor which would result in a performance similar to a non-convolutional recurrent neural network.
Having the unary constraints in the network can make the other constraints perform better as well, as they can correlate with the behaviour of the presence/absence of particular activities.

The impact of particular constraints is covered as well. 
Unary constraints perform well in most cases, and binary constraints' performance is strongly linked to their presence.
Looking back to our motivation in Section \ref{sec:moti}, it appears that it is easier to predict the occurrence of unary relationships with high precision, making it possible to, e.g., verify whether an application is going to be rejected (depending on whether \textit{absence} is predicted to hold for A\_Declined), rather than whether a particular activity precedes or always follows another (e.g. the chain response constraint between A\_Declined and W\_Completeren aanvraag).
Nevertheless, high precision is still achieved at +80\% and +90\% levels for BPI 17.
Hence, questions such as: `are all final A\_Declined preceded by W\_Completeren aanvraag' can be predicted well using PAM similar to the objectives predicted in \cite{DBLP:journals/corr/TeinemaaDRM17}.
However, PAM predicts over 9,000 constraints/objectives simultaneously (e.g. for 26 activities over 14 binary constraints for BPI 17), although they are not as tailored towards a particular question and overlap to some extent (e.g. \textit{exactly(a)} overlaps with \textit{exactly(a,2)} and \textit{precedence(a,b)}).
PAM allows to combine various output constraints to create even stronger objective sets or even declarative process models, but does not include any data variables within the LTL constraints.
Finally, PAM is not as flexible regarding the input, where only a final window is predicted and the window size or number of windows is determined up front. 
This might require multiple runs to finetune results and depends on the outcome of the analysis.
E.g., in some cases a fixed window length might be preferred over predicting the last window, as it might not be possible to determine how many more windows will be present.
Still, PAM can be run regardless of this knowledge and provide a prediction by dividing the current execution trace in a fixed number of windows, or windows of a fixed length.

\section{Conclusion and Future Work}
\label{sec:conc}
This paper introduced PAM, a new technique that encompasses a feature generation approach for processes such that processes can be treated as movies fit for training high-dimensional recurrent neural networks.
This allows modelling the dynamic development of process models, offering a mix of next-in-sequence and objective-based predictive process modelling, as well as process model forecasting.
It is shown that declarative process constraints can be used to make a multi-dimensional representation of activity pairs to capture a process model at various points in time, which is used towards predicting the constraint set constituting the process model that will be present in subsequent windows of the process execution.
The neural network architecture has been proven to be adequate to perform this prediction, with high accuracy and precision, for various real-life event logs and different approaches to extract windows from traces.
Hence, this makes PAM effective to provide a forecast and early warning of constraint violations and satisfaction, allowing the support of monitoring process behaviour and even predictive conformance checking.

There are many future directions for this research.
Firstly, a wider experimental evaluation with an even larger hyperparameter search, focusing on deeper or different network architectures such as PredRNN \cite{DBLP:conf/icml/WangGLWY18} can be investigated.
Next, a more in-depth analysis of various constraint types would be worthwhile.
Some constraint types might be more interesting, which results in a trade-off between depth and performance.
If constraints are seldomly present, they might not provide sufficient learning material for the network, while they increase the dimensionality of the input.
Hence, it might be sufficient to use PAM with only a subset of constraints to obtain similar insights and results.
The window-based approach can be further improved significantly as well.
Currently, the setup of splitting up traces in windows of equal length or in an equal number of windows could be improved in a variety of ways to account for longer and shorter traces, and to find anchor points reflecting particular milestones within a process which might be better suitable for creating models that are self-contained.
Finally, it would be interesting to see whether the technique can run in an on-line fashion, by pre-training a network and updating it according to newly-generated traces.

\bibliographystyle{abbrvnat}

\end{document}